\documentclass{article}

\usepackage{microtype}
\usepackage{graphicx}
\usepackage{booktabs} % for professional tables
\usepackage{amsthm, amsfonts, amsmath}
\usepackage{subcaption}

\newtheorem{property}{Property}
\newcommand{\minisection}[1]{\vspace{2mm}\noindent{\textbf{#1}.}}

\usepackage{hyperref}
\usepackage{natbib}

\usepackage[accepted]{icml2020}

\icmltitlerunning{LEEP: A New Measure to Evaluate Transferability of Learned Representations}

\begin{document}

\twocolumn[
\icmltitle{LEEP: A New Measure to Evaluate Transferability of Learned Representations}

% List of affiliations: The first argument should be a (short)
% identifier you will use later to specify author affiliations
% Academic affiliations should list Department, University, City, Region, Country
% Industry affiliations should list Company, City, Region, Country

\begin{icmlauthorlist}
\icmlauthor{Cuong V.~Nguyen}{aws}
\icmlauthor{Tal Hassner}{fb}
\icmlauthor{Matthias Seeger}{aws}
\icmlauthor{Cedric Archambeau}{aws}
\end{icmlauthorlist}

\icmlaffiliation{aws}{Amazon Web Services}
\icmlaffiliation{fb}{Facebook AI (Work done before joining Facebook)}

\icmlcorrespondingauthor{Cuong V.~Nguyen}{nguycuo@amazon.com}

% You may provide any keywords that you
% find helpful for describing your paper; these are used to populate
% the "keywords" metadata in the PDF but will not be shown in the document
\icmlkeywords{transfer learning, transferability, deep learning}

\vskip 0.3in
]

\printAffiliationsAndNotice{}  % leave blank if no need to mention equal contribution

\begin{abstract}
We introduce a new measure to evaluate the transferability of representations learned by classifiers. Our measure, the \emph{Log Expected Empirical Prediction} (LEEP), is simple and easy to compute: when given a classifier trained on a source data set, it only requires running the target data set through this classifier once. We analyze the properties of LEEP theoretically and demonstrate its effectiveness empirically. Our analysis shows that LEEP can predict the performance and convergence speed of both transfer and meta-transfer learning methods, even for small or imbalanced data. Moreover, LEEP outperforms recently proposed transferability measures such as negative conditional entropy and H scores. Notably, when transferring from ImageNet to CIFAR100, LEEP can achieve up to 30\% improvement compared to the best competing method in terms of the correlations with actual transfer accuracy.
\end{abstract}

\section{Introduction}
\label{sec:intro}

Transferability estimation~\citep{eaton2008modeling, ammar2014automated, sinapov2015learning} is the problem of quantitatively estimating how easy it is to transfer knowledge learned from one classification task to another. Specifically, given a source task, represented by a labeled data set or a pre-trained model, and a target task, represented by a labeled data set, transferability estimation aims to develop a measure (or a score) that can tell us, ideally without training on the target task, how effectively transfer learning algorithms can transfer knowledge from the source task to the target task.

Answering this question is important, since good estimations of transferability can help understand the relationships between tasks~\citep{tran2019transferability}, select groups of highly transferable tasks for joint training~\citep{zamir2018taskonomy}, or choose good source models for a given target task~\citep{achille2019task2vec, bao2019information, bhattacharjee2019p2l}. Previous approaches to transferability estimation often require running a transfer learning algorithm that involves expensive parameter optimization~\citep{zamir2018taskonomy, achille2019task2vec}, do not have a simple interpretation~\citep{bao2019information}, or make strong assumptions about the data sets that limit their applicability~\citep{zamir2018taskonomy, tran2019transferability}.

We propose a novel measure called the \emph{\textbf{L}og \textbf{E}xpected \textbf{E}mpirical \textbf{P}rediction} (LEEP) for transferability estimation of deep networks that overcomes all the shortcomings above. In contrast to previous approaches, LEEP scores are obtained without training on the target task, thus avoiding the expensive parameter optimization step. Additionally, they have a simple interpretation and can be applied in general settings to a wide range of modern deep networks.

In particular, LEEP scores are obtained from a source model and a target data set by making a \emph{single} forward pass of the model through the target data. This is a simpler process than previous methods, such as Taskonomy~\citep{zamir2018taskonomy} and Task2Vec~\citep{achille2019task2vec}, where one has to re-train at least part of the source model on the target data set. Furthermore, LEEP has a simple interpretation: it is the average log-likelihood of the \emph{expected empirical predictor}, a simple classifier that makes prediction based on the expected empirical conditional distribution between source and target labels. Finally, LEEP does not make any assumption on the source and target input samples, except that they have the same size. This is more general and applicable than previous work~\citep{zamir2018taskonomy, tran2019transferability} where source and target data sets were assumed to share the same input samples.

\minisection{Contributions}
We formally define LEEP and rigorously analyze it, both theoretically and empirically. We show two theoretical properties of the measure: (1) LEEP is upper bounded by the average log-likelihood of the optimal model, obtained by re-training the head classifier while freezing the feature extractor; (2) LEEP is related to the negative conditional entropy measure proposed by~\citet{tran2019transferability}.

We conduct extensive experiments to evaluate our LEEP measure in several scenarios. We show that the measure is useful for predicting the performance of two commonly used transfer learning algorithms -- head classifier re-training~\citep{donahue2014decaf, sharif2014cnn} and model fine-tuning~\citep{agrawal2014analyzing, girshick2014rich} -- not only for large target data sets, but also for small or imbalanced target data sets that are difficult to use for re-training. We also show that LEEP can predict the convergence speed of the fine-tuning method for transfer learning. 

We further demonstrate that LEEP can predict the performance of a recently developed meta-transfer learning method, the Conditional Neural Adaptive Processes~\citep{requeima2019fast}. Meta-transfer learning~\citep{ying2018transfer, sun2019meta, requeima2019fast} is a framework for learning to transfer using several meta-training tasks. Importantly, to our knowledge, our work is the first to develop a transferability measure for meta-transfer learning.

We empirically compare our method with the very recent negative conditional entropy measure~\citep{tran2019transferability} and H scores~\citep{bao2019information}. Our comparisons show that LEEP better correlates with the actual transfer accuracy than these methods. Finally, we demonstrate the effectiveness of LEEP for the source model selection problem in comparison with the negative conditional entropy and H scores.

\section{Log Expected Empirical Prediction}
\label{sec:leep}

Consider transfer learning between two classification tasks: a source task, represented by a pre-trained model, and a target task, represented by a labeled data set. Formally, assume the source task requires learning a model that maps input instances from a domain $\mathcal{X} = \mathbb{R}^N$ to labels in a finite label set $\mathcal{Z}$. Further, assume that we already trained such a model, which we call the \emph{source model}, denoted by $\theta$. We note that $\mathcal{X}$ can contain text or images since they can be flatten to vectors in $\mathbb{R}^N$. Transfer learning seeks to learn a model for a target task mapping inputs from $\mathcal{X}$ to some finite target label set $\mathcal{Y}$, given a \emph{target data set} $\mathcal{D} = \{ (x_1, y_1), (x_2, y_2), \ldots, (x_n, y_n) \}$, where $x_i \in \mathcal{X}$ and $y_i \in \mathcal{Y}$, for this purpose. We emphasize that, unlike recent previous work~\citep{zamir2018taskonomy, tran2019transferability}, the domain $\mathcal{X}$ in our setting is general: the input instances of the source and target tasks are only assumed to have the same dimension,\footnote{If the source model is fully convolutional~\cite{long2015fully}, this assumption can be relaxed.} such as ImageNet~\citep{russakovsky2015imagenet} and CIFAR~\citep{krizhevsky2009learning} images scaled to the same size.

The transfer learning setting above is very common in computer vision for leveraging expensive pre-trained deep learning models. For instance, a popular transfer learning method takes a model, pre-trained on ImageNet, and re-trains its classifier (the head) on the target data set while freezing the feature extractor. The result is a new model for the target task that uses the representation learned on the source with a head classifier learned on the target~\citep{donahue2014decaf, sharif2014cnn, zeiler2014visualizing, oquab2014learning, whatmough2019fixynn}. We call this the \emph{head re-training} method. Another popular transfer learning method, called the \emph{fine-tuning} method, fine-tunes the feature extractor while re-training the new head classifier on the target data set to get a model for the target task~\citep{agrawal2014analyzing, girshick2014rich, chatfield2014return, dhillon2020baseline}.

In this work, we study the \emph{transferability estimation} problem, which aims to develop a measure (or a score) that can tell us, without training on the target data set, how effectively these transfer learning algorithms can transfer knowledge learned in the source model $\theta$ to the target task, using the target data set $\mathcal{D}$. We emphasize that although transferability estimation can tell us the effectiveness of transfer learning, it is not a solution for transfer learning in itself. Instead, transfer learning is achieved by e.g., the head re-training or fine-tuning methods above. In our experiments in Sec.~\ref{sec:experiments}, we will use these transfer learning methods to test our transferability measure.

We now describe our proposed measure, LEEP, that requires no expensive training on the target task and offers an \emph{a priori} estimate of how well a model will transfer to the target task. The measure can be efficiently computed from the source model $\theta$ and the target data set $\mathcal{D}$. The computation involves the following three steps.

\vspace{-1.5mm}
\minisection{Step 1: Compute dummy label distributions of the inputs in the target data set $\mathcal{D}$}
We apply $\theta$ to each input $x_i$ in $\mathcal{D}$ to get the predicted distribution over $\mathcal{Z}$, the label set of the source task. We denote this predicted distribution by~$\theta(x_i)$, which is a categorical distribution over $\mathcal{Z}$. Note that $\theta(x_i)$ is a \emph{dummy distribution} over labels of the source task, since these labels may not be meaningful for the example $x_i$. For instance, if~$\theta$ is a model pre-trained on ImageNet and $\mathcal{D}$ is the CIFAR data set, then $\theta(x_i)$ is a distribution over ImageNet labels, which may not be semantically related to the true label of~$x_i$ in the CIFAR data set.

\vspace{-1.5mm}
\minisection{Step 2: Compute the empirical conditional distribution $\hat{P}(y | z)$ of the target label $y$ given the source label $z$}
We next compute the empirical conditional distribution $\hat{P}(y | z)$ for all $(y, z) \in \mathcal{Y} \times \mathcal{Z}$. To this end, we first compute the empirical joint distribution $\hat{P}(y, z)$ for all label pairs $(y, z) \in \mathcal{Y} \times \mathcal{Z}$:
\begin{equation}
\hat{P}(y, z) = \frac{1}{n} \sum_{i:y_i=y} \theta(x_i)_z,
\end{equation}
where the summation $\sum_{i:y_i=y}$ means we sum over all indices $i \in \{ 1, 2, \ldots, n\}$ that satisfy $y_i = y$. In the above equation, $\theta(x_i)_z$ is the probability of the label $z$ according to the categorical distribution $\theta(x_i)$.

From this empirical joint distribution, we can compute the empirical marginal distribution $\hat{P}(z)$:
\begin{align*}
\hat{P}(z) &= \sum_{y \in \mathcal{Y}} \hat{P}(y, z) = \frac{1}{n} \sum_{i=1}^n \theta(x_i)_z,
\end{align*}
and then the empirical conditional distribution $\hat{P}(y | z)$:
\begin{align*}
\hat{P}(y | z) &= \frac{\hat{P}(y, z)}{\hat{P}(z)}.
\end{align*}

\vspace{-5mm}
\minisection{Step 3: Compute LEEP using $\theta(x)$ and $\hat{P}(y | z)$}
For any input $x \in \mathcal{X}$, consider a classifier that predicts a label $y$ of $x$ by first randomly drawing a dummy label $z$ from $\theta(x)$ and then randomly drawing $y$ from $\hat{P}(y | z)$. Equivalently, this classifier can predict $y$ by directly drawing a label from the distribution $p(y | x; \theta, \mathcal{D}) = \sum_{z \in \mathcal{Z}} \hat{P}(y | z) ~ \theta(x)_z$. We call this classifier the \emph{\textbf{E}xpected \textbf{E}mpirical \textbf{P}redictor} (EEP).

For the target data set $\mathcal{D}$, we define LEEP as the average log-likelihood of the EEP classifier given the data $\mathcal{D}$. Formally, our LEEP measure of transferability is defined as:
\begin{equation}
T(\theta, \mathcal{D}) = \frac{1}{n} \sum_{i=1}^n \log \left( \sum_{z \in \mathcal{Z}} \hat{P}(y_i | z) ~ \theta(x_i)_z \right).
\label{eq:leep}
\end{equation}
Intuitively, this measure tells us how well the EEP performs on $\mathcal{D}$. We argue that since the EEP is constructed mainly from the source model $\theta$ with a minimal use of $\mathcal{D}$ (i.e., $\mathcal{D}$ is only used to compute the simple empirical conditional distribution), it can serve as an indicator of how ``close'' $\theta$ and $\mathcal{D}$ are, and hence a measure of transferability.

From its definition, the LEEP measure is always negative and larger values (i.e., smaller absolute values) indicate better transferability. When the target task contains more classes, LEEP scores tend to be smaller. The measure is also efficient to compute since its computational bottleneck (step 1 above) requires only a single forward pass through the target data set $\mathcal{D}$.

\vspace{-1mm}
\section{Theoretical Properties of LEEP}
\label{sec:theory}
\vspace{-1mm}

Assume $\theta = (w, h)$ where $w$ is a feature extractor that maps an input $x \in \mathcal{X}$ to a representation (or \emph{embedding}), $r = w(x)$, and $h$ is a classifier (or \emph{head}) that takes the representation $r$ as input and returns a probability distribution over $\mathcal{Z}$. Next, assume we fix $w$ and re-train the classifier by maximum likelihood, using the target data set $\mathcal{D}$, to obtain the new classifier $k^*$. That is,
\begin{equation}
k^* = \arg \max_{k \in \mathcal{K}} ~ l(w, k),
\label{eq:opt_loglik}
\end{equation}
where $l(w, k) = \frac{1}{n} \sum_{i=1}^n \log p(y_i | x_i; w, k)$ is the average log-likelihood of $(w, k)$ on the target data set $\mathcal{D}$, and $\mathcal{K}$ is the space of classifiers from which we would like to select $k$. We note that $\mathcal{K}$ contains classifiers that map $w(x)$ to labels in $\mathcal{Y}$. If we assume $\mathcal{K}$ contains the EEP, we can easily show the following property of the LEEP measure (see proof in Appendix~\ref{sec:proofs}):
\begin{property}
LEEP is a lower bound of the optimal average log-likelihood. Formally, $T(\theta, \mathcal{D}) \le l(w, k^*)$.
\label{prop:upper_bound}
\end{property}
The assumption that $\mathcal{K}$ contains the EEP can be easily satisfied if we select $\mathcal{K} = \bar{\mathcal{K}} \cup \{ k_{\text{EEP}} \}$, where the classifier $k_{\text{EEP}}$ is the EEP and $\bar{\mathcal{K}}$ is a space of classifiers that we can easily optimize over (e.g., the parameter space of a linear classifier). We can then solve Eq.~\eqref{eq:opt_loglik} by the following two-stage process. First, we solve $\bar{k} = \arg \max_{k \in \bar{\mathcal{K}}} ~ l(w, k)$ using, for instance, stochastic gradient descent (SGD)~\citep{robbins1951stochastic, bottou1991stochastic}. Then, we solve $k^* = \arg \max_{k \in \{ \bar{k}, k_{\text{EEP}} \}} l(w, k)$ by simple comparison. In our experiments, we skip this second step and choose $k^* = \bar{k}$, as usually done in practice. The results reported in Sec.~\ref{sec:predict_exp} show that LEEP correlates well with the test accuracy of $\bar{k}$, indicating that the measure can be used to estimate the performance of the transferred model in practice.

As a second property, we show a relationship between LEEP and the negative conditional entropy (NCE) measure recently proposed by~\citet{tran2019transferability}. To measure the transferability between $\theta$ and $\mathcal{D}$, we can compute the NCE measure as follows. First, we use $\theta$ to label all $x_i$'s in $\mathcal{D}$. For example, we can label $x_i$ by a dummy label $z_i = \arg \max_{z \in \mathcal{Z}} \theta(x_i)_z$. Then, using the dummy label set $Z = (z_1, z_2, \ldots, z_n)$ and the true label set $Y = (y_1, y_2, \ldots, y_n)$, we can compute $\mathrm{NCE}(Y | Z)$. It can be shown that the following property relating LEEP and NCE holds (see proof in Appendix~\ref{sec:proofs}):
\begin{property}
LEEP is an upper bound of the NCE measure plus the average log-likelihood of the dummy labels. Formally, $T(\theta, \mathcal{D}) \ge \mathrm{NCE}(Y | Z) + \frac{1}{n} \sum_{i=1}^n \log \theta(x_i)_{z_i}$.
\label{prop:lower_bound}
\end{property}
From Properties~\ref{prop:upper_bound} and~\ref{prop:lower_bound}, we know that LEEP is bounded between $\mathrm{NCE}(Y | Z) + \frac{1}{n} \sum_i \log \theta(x_i)_{z_i}$ and the average log-likelihood of the re-trained model. When the re-trained model does not overfit, its average log-likelihood is a reasonable indicator of the model's performance~\citep{tran2019transferability}. When LEEP is close to this average log-likelihood, it can be considered, in some sense, a reasonable indicator of transferability. Our experiments in Sec.~\ref{sec:compare_exp} show the effectiveness of LEEP over NCE. 

We note that in the setting of \citet{tran2019transferability}, the source data set is given, instead of the source model $\theta$. Hence, NCE is a more natural measure of transferability in their case. In contrast, LEEP is more natural for our setting, since we are only given the source model $\theta$. In return, by assuming a source model, LEEP is not restricted to their setting, where the two tasks are defined over the exact same input instances.

\section{Related Work}
\label{sec:related_work}

Our work is related to several research areas in machine learning and computer vision, including transfer learning~\citep{weiss2016survey, yang2020transfer}, meta-transfer learning~\citep{ying2018transfer, sun2019meta, requeima2019fast}, task space modeling~\citep{zamir2018taskonomy, achille2019task2vec}, and domain adaptation~\citep{sun2016return, azizzadenesheli2019regularized}. We discuss below previous work that is closely related to ours.

\minisection{Transfer learning}
Our paper addresses the problem of predicting the performance of transfer learning algorithms between two classification tasks, without actually executing these algorithms. This problem is also called transferability estimation between classification tasks~\citep{bao2019information, tran2019transferability}. Early theoretical work in transfer and multi-task learning studied the \emph{relatedness} between tasks and proposed several types of distances between tasks. These  distances include the $\mathcal{F}$-relatedness~\citep{ben2003exploiting}, $\mathcal{A}$-distance~\citep{kifer2004detecting, ben2007analysis}, and discrepancy distance~\citep{mansour2009domain}. Although useful for theoretical analysis, these approaches are unsuited for measuring transferability in practice because they cannot be computed easily and are symmetric. Transferability measures should be non-symmetric since transferring from one task to another (e.g., from a hard task to an easy one) is different from transferring in the reverse direction (e.g., from the easy task to the hard one).

More recently, \citet{azizzadenesheli2019regularized} studied domain adaptation under label shift. The authors considered the case where only the marginal label distribution changes between domains. This assumption is more restrictive than the setting in our paper as we allow both the input distribution and the label distribution to change arbitrarily. Task similarity was also considered for Gaussian process based transfer learning between regression tasks~\citep{cao2010adaptive, wei2018uncluttered}.

The most related work to ours are transferability measures recently proposed by~\citet{tran2019transferability} and~\citet{bao2019information}.~\citet{tran2019transferability} developed the negative conditional entropy measure between the source and target label sets, under the assumption that the source and target data sets share the same input examples. Our paper removes this requirement and allows the input data to come from arbitrarily different distributions.~\citet{bao2019information} developed a transferability measure based on H scores, which are derived from information-theoretic principles. Although, like us, their transferability measure can be applied to general settings, their measure is hard to interpret since it involves solving a Hirschfeld-Gebelein-R\'enyi maximum correlation problem~\citep{hirschfeld1935connection, gebelein1941statistische, renyi1959measures}. LEEP scores, on the other hand, have a simple interpretation related to the EEP and are easy to implement and compute.

\minisection{Meta-transfer learning}
Meta-transfer learning is a framework for learning to transfer from a source task to a target task~\citep{ying2018transfer, sun2019meta, requeima2019fast}. Similar to our transfer learning setting,~\citet{sun2019meta} and~\citet{requeima2019fast} also adapted a pre-trained model on the source task to the target task. These meta-learning methods learn the adaptation from several meta-training tasks, which consist of additional target data sets and target test sets from different domains that are intended to mimic the transfer learning scenario, where one wants to transfer knowledge to \emph{unseen} tasks. Because of the additional data sets, the transfer learning mechanism in meta-learning departs from that of regular transfer learning. Nevertheless, we show that LEEP scores can also predict the performance of meta-transfer learning algorithms, such as the conditional neural adaptive processes (CNAPs) recently proposed by~\citet{requeima2019fast}. To our knowledge, we are the first to develop a transferability measure that can be applied to meta-transfer learning.

\minisection{Task space representation}
Our paper is related to task space representation~\citep{edwards2016towards, zamir2018taskonomy, achille2019task2vec, jomaa2019dataset2vec} in the sense that transferability may be estimated from a distance between the representations (or embeddings) of tasks. For instance, Task2Vec~\citep{achille2019task2vec} tried to map tasks (or data sets) to vectors in a vector space. Transferability between tasks was then estimated using a non-symmetric distance between the corresponding vectors. This method requires training a large reference network (called the probe network), adapting it to the target data set, and computing the Fisher information matrix to obtain a task embedding. Our method, on the other hand, is much simpler and computationally more efficient. Additionally, we require only the source model and a small target data set, which are both usually available in practice.

\citet{edwards2016towards} extended the variational auto-encoder~\citep{kingma2014stochastic} to compute statistics of data sets that are useful in a range of applications, including clustering data sets or classifying unseen classes. These statistics, however, were not shown to be useful for transferability estimation. Another related work, Taskonomy~\citep{zamir2018taskonomy}, created a taxonomy of tasks that revealed task structures useful for reducing the number of labeled training data. Taskonomy involves several steps, one of which requires computing the task affinity matrix by re-training networks on target data sets. LEEP scores can be regarded as an efficient approximation of the task affinity, as we avoid network re-training.

\section{Experiments}
\label{sec:experiments}

We evaluate the ability of LEEP to predict the performance of transfer and meta-transfer learning algorithms, prior to applying these algorithms in practice. We further show that LEEP is useful even in the small or imbalanced data settings, where training on the target task could be hard. We compare LEEP with the state of the art NCE transferability measure of~\citet{tran2019transferability} and H score of~\citet{bao2019information}. Finally, we demonstrate the use of LEEP for source model selection. Our experiments are implemented in Gluon/MXNet~\citep{chen2015mxnet, guo2019gluoncv}.

\subsection{LEEP vs.~Transfer Accuracy}
\label{sec:predict_exp}

We show that LEEP scores effectively predict the accuracies of models transferred from source to target tasks. We consider two different source models, each one representing a different source task: ResNet18~\citep{he2016deep}, which is pre-trained on ImageNet~\citep{russakovsky2015imagenet}, and ResNet20~\citep{he2016deep}, which is pre-trained on CIFAR10~\citep{krizhevsky2009learning}. For each model, we construct 200 different target tasks from the CIFAR100 data set~\citep{krizhevsky2009learning} as follows. The label set of each target task is constructed by randomly drawing a subset of the 100 classes, with the subset's size ranging from 2 to 100. The target data set then consists of all training examples of the selected classes in the CIFAR100 data set. We use the test examples of these selected classes as the target test set to compute the accuracy of the transferred model.

We experiment with two commonly used transfer learning methods to train a transferred model to a target task:

\minisection{Re-train head} This method keeps the feature extractor layers of the source model fixed and then trains a new head classifier using the target data set from scratch. We re-train the new classifier by running SGD on the cross entropy loss.

\minisection{Fine-tune} This method replaces the head classifier of the source model with a new head and then fine-tunes the entire model -- the feature extractor and the new head -- using the target data set. Fine-tuning is performed again by running SGD on the cross entropy loss.

To clarify, we let the feature extractor be the portion of the source model's network up to and including the penultimate layer. The head classifier is the network' last fully connected layer. For each target task, models transferred using these two methods were evaluated on the target test set to obtain the test accuracies. We then compare these accuracies with our LEEP scores evaluated on these target tasks. 

In all tests, we ran SGD for 100 epochs with learning rate 0.01 and batch size 10 since they were sufficient to obtain good transferred models. We found that varying the number of epochs does not significantly change the results of our experiments, although in principle, fine-tuning the whole network until convergence could decouple its dependence on the source task.

\begin{figure}[t]
\begin{center}
\centerline{
\begin{subfigure}[t]{0.23\textwidth}
\includegraphics[width=\textwidth]{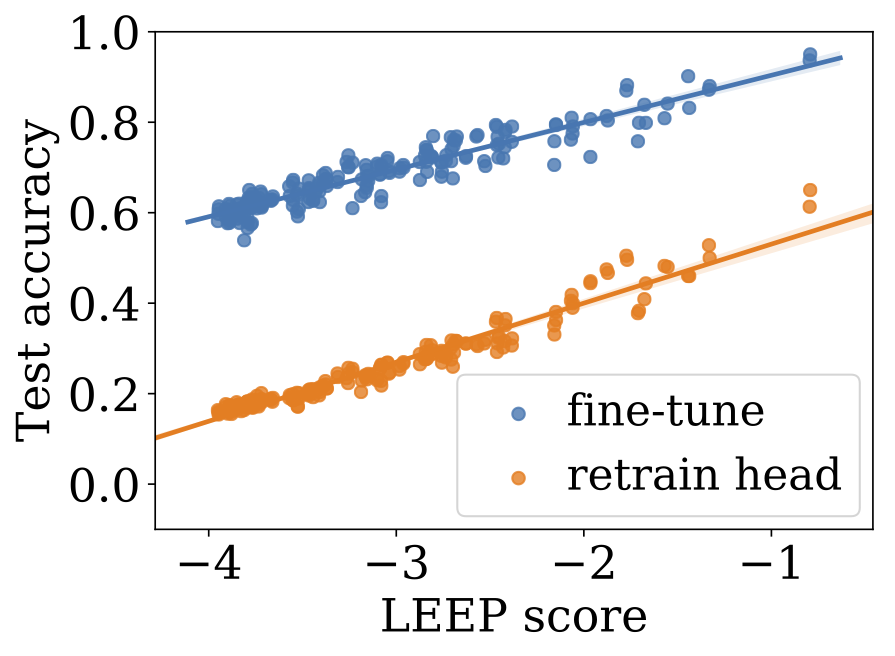}
\caption{ImageNet $\rightarrow$ CIFAR100}
\end{subfigure}
\hspace{1.5mm}
\begin{subfigure}[t]{0.23\textwidth}
\includegraphics[width=\textwidth]{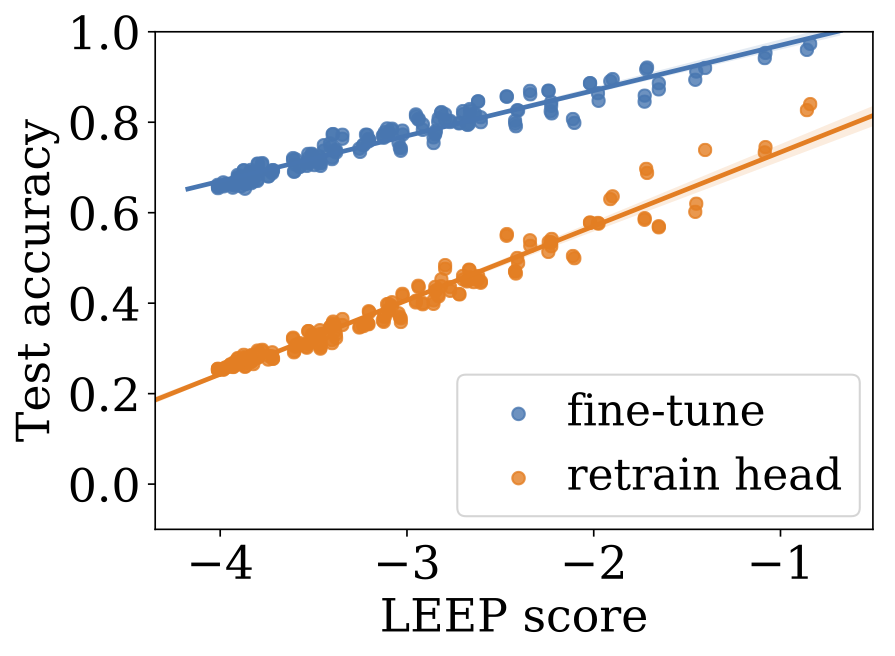}
\caption{CIFAR10 $\rightarrow$ CIFAR100}
\end{subfigure}
}
\caption{{\bf LEEP scores vs.~test accuracies} of two transfer learning algorithms, together with their best fit lines, reported for transferred models on 200 random tasks constructed from CIFAR100 data. Blue and orange points on a vertical line correspond to the same target data set. The source models are (a) ResNet18 pre-trained on ImageNet, and (b) ResNet20 pre-trained on CIFAR10. See Sec.~\ref{sec:predict_exp} for details.}
\label{fig:predict}
\end{center}
\vskip -0.2in
\end{figure}

Fig.~\ref{fig:predict} shows the LEEP scores and the corresponding test accuracies for transferred models on 200 target tasks. Following the correlation analysis by \citet{nguyen2019toward} and \citet{tran2019transferability}, we compute the Pearson correlation coefficients and the $p$ values between LEEP scores and test accuracies, which are shown in Table~\ref{tab:leep_compare} (see the first two rows of each transfer method). From Fig.~\ref{fig:predict} and Table~\ref{tab:leep_compare}, LEEP scores clearly correlate with test accuracies, with correlation coefficients higher than 0.94 and $p<0.001$ in all cases. Evidently, LEEP scores are a reliable indicator of the performance of transferred models.

\subsection{LEEP vs.~Transfer Accuracy in Small Data Regime}
\label{sec:small_data}

Transfer learning is often performed in cases where target data sets are small. We next evaluate LEEP in such small data settings. We focus on scenarios where the size of the target data set might be insufficient to train a deep model from scratch, but reasonable for transfer learning to be effective. Particularly, we repeat the experiments of Sec.~\ref{sec:predict_exp} with the additional restriction that target data sets contain exactly five random classes and 50 random examples per class. Thus, the target data sets contain only 10\% of the training examples per class, compared to the full CIFAR100 data set.

We further add experiments where target data sets are constructed from the FashionMNIST data set~\citep{xiao2017fashion}. In this case, the target data sets are restricted to contain four random classes and 30 random examples per class (i.e., 0.5\% of the training examples per class compared to the full data set). Since FashionMNIST contains grayscale images, we duplicate the grayscale channel three times to be consistent with the source model's input domain. Because the target data set size is small, we re-train the head classifier by running SGD for only 50 epochs.

Table~\ref{tab:leep_compare} (see the 3$^{\text{rd}}$ to 6$^{\text{th}}$ rows of the first two algorithms) shows the correlation coefficients between LEEP scores and test accuracies in these settings. The results indicate a reasonably positive correlations with coefficients greater than 0.5 in most cases, with the exception of fine-tuning from the CIFAR10 pre-trained model. We note that, in general, the number of examples per class in the target data set would affect the effectiveness of LEEP. For instance, if the number of examples is too small, the empirical conditional distribution, and consequently LEEP, would become unreliable.

\begin{figure}[t]
\begin{center}
\centerline{
\begin{subfigure}[t]{0.235\textwidth}
\includegraphics[width=\textwidth]{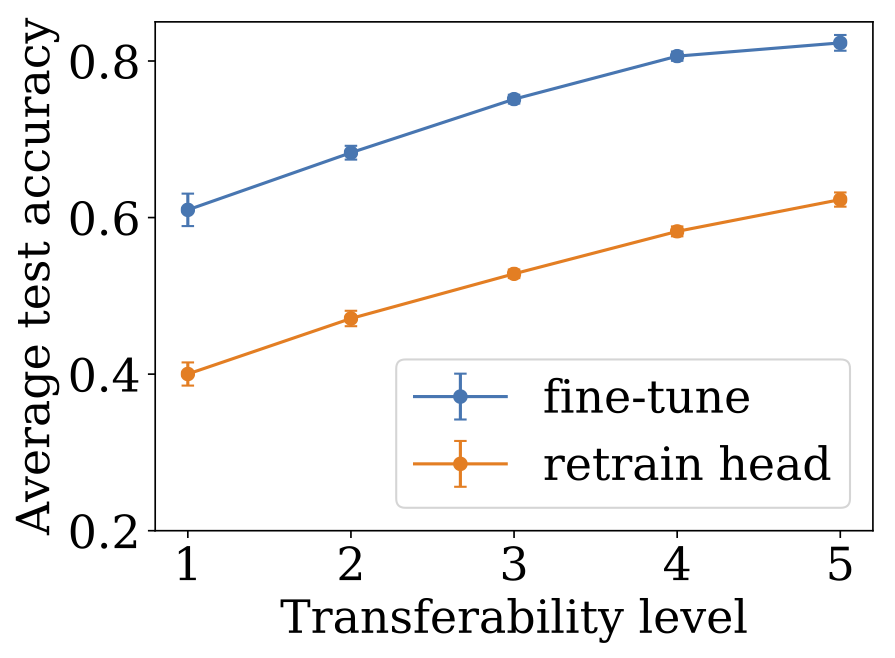}
\caption{Small balanced target data}
\end{subfigure}
\hspace{1.5mm}
\begin{subfigure}[t]{0.235\textwidth}
\includegraphics[width=\textwidth]{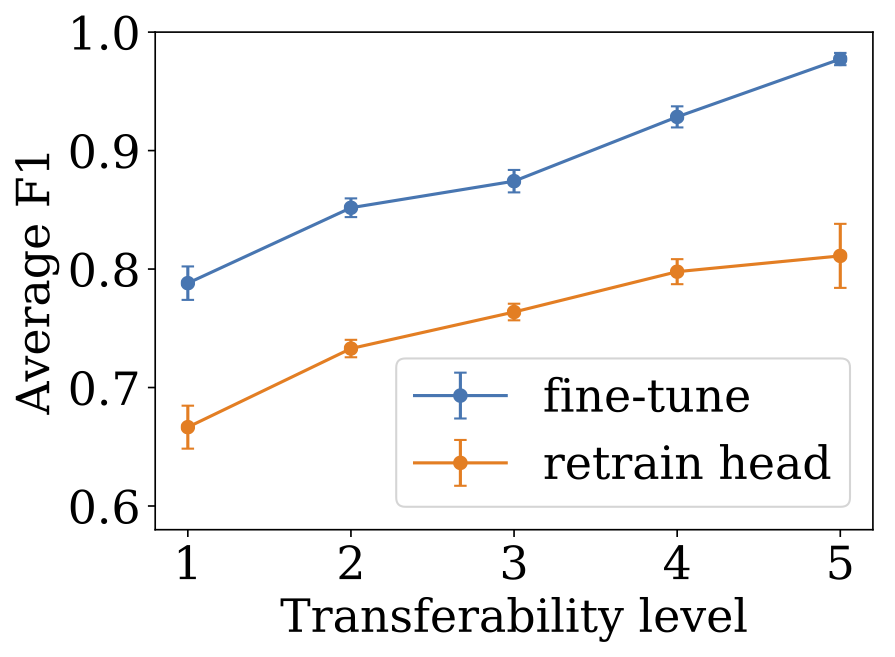}
\caption{Small imbalanced target data}
\end{subfigure}
}
\caption{{\bf Performance of transferred models on small target data sets in five transferability levels} predicted from LEEP scores. The higher the level, the easier the transfer. The transfer is from a ResNet18 pre-trained on ImageNet to target tasks constructed from CIFAR100. We considered both (a) balanced and (b) imbalanced target data sets. See Sec.~\ref{sec:small_data} and~\ref{sec:skewed_data} for details.}
\label{fig:predict_small}
\end{center}
\vskip -0.25in
\end{figure}

To factor out the noise when evaluating LEEP scores on small target data sets, we also consider partitioning the scores' range into five equal bins and averaging the test accuracies of tasks in each bin. This allows us to compare the average test accuracies of tasks belonging in five transferability levels. Fig.~\ref{fig:predict_small}(a) shows a typical result for this case, where higher transferability levels generally imply better accuracies for both transfer methods. Full results are given in Fig.~\ref{fig:predict_small_full} in Appendix~\ref{sec:full_results}. These results testify that LEEP scores can predict accuracies of transferred models, even in small data regimes.

We also evaluate LEEP in the noisy, small data setting, where we repeat the experiment between ImageNet and CIFAR100 but randomly flip the labels in the target data set to a wrong one with probability 0.15. We report the correlation coefficients for this case in Table~\ref{tab:leep_compare} (see the 7$^{\text{th}}$ rows of the first two algorithms). The results also show positive correlations between LEEP scores and test accuracies, although the correlations are weaker than those with correct labels.

\subsection{LEEP vs.~F1 Score on Imbalanced Data}
\label{sec:skewed_data}

Imbalanced target data sets are commonly encountered in practice~\citep{al2016transfer, wang2017balanced}. We now evaluate LEEP in this setting. Specifically, we also repeat the experiments of Sec.~\ref{sec:predict_exp} and~\ref{sec:small_data} where, this time, we restrict target data sets to imbalanced binary classification sets. The two classes of each target data set are drawn randomly. The size of the smaller class is chosen uniformly from 30 to 60, while the other class is five times larger. Because the target data sets are small with only two classes, we only need to re-train the head or fine-tune for 20 epochs.

Table~\ref{tab:leep_compare} (see the last four rows of the first two algorithms) shows the correlation coefficients between LEEP scores and test F1 scores in these settings. The results also indicate a reasonably positive correlations with coefficients greater than 0.5 in all cases. Fig.~\ref{fig:predict_small}(b) further shows the average test F1 scores in five transferability levels predicted from LEEP for a typical case, where higher transferability levels imply better F1 scores for both transfer methods. Full results are given in Fig.~\ref{fig:predict_skewed_full} in Appendix~\ref{sec:full_results}. These results again confirm that LEEP scores can predict the performance of transferred models, even for imbalanced target data sets.

\subsection{LEEP vs.~Accuracy of Meta-Transferred Models}
\label{sec:meta_transfer_exp}

Meta-transfer learning is a framework for learning to adapt from a source task to a target task~\citep{ying2018transfer, sun2019meta, requeima2019fast}. Next, we show that LEEP can also predict the test accuracies of CNAPs~\citep{requeima2019fast}, a recently proposed meta-transfer learning method. CNAPs adapt a pre-trained model on the source task to a target task by adding scale and shift parameters to each channel of its convolutional layers. The additional parameters are outputs of adaptation networks, which are trained by meta-learning from several training tasks. When given a target data set, the adaptation networks return the appropriate scale and shift parameters that can augment the source model to make predictions on the target test set.

In our experiment, we follow the original training procedure proposed for CNAPs by~\citet{requeima2019fast}, where the source model is a ResNet18 pre-trained on ImageNet, and the adaptation networks are trained using the Meta-data set~\citep{triantafillou2019meta}. We also test CNAPs on 200 random target tasks drawn from CIFAR100 as follows. Each target data set contains five random labels and 50 random examples per class, drawn from the test set of CIFAR100. The remaining 50 test examples of the selected classes are used as the target test set. This testing procedure is consistent with the one used by~\citet{requeima2019fast}, except that in our experiments, we fix the number of classes.

\begin{figure}[t]
\begin{center}
\centerline{
\includegraphics[width=0.3\textwidth]{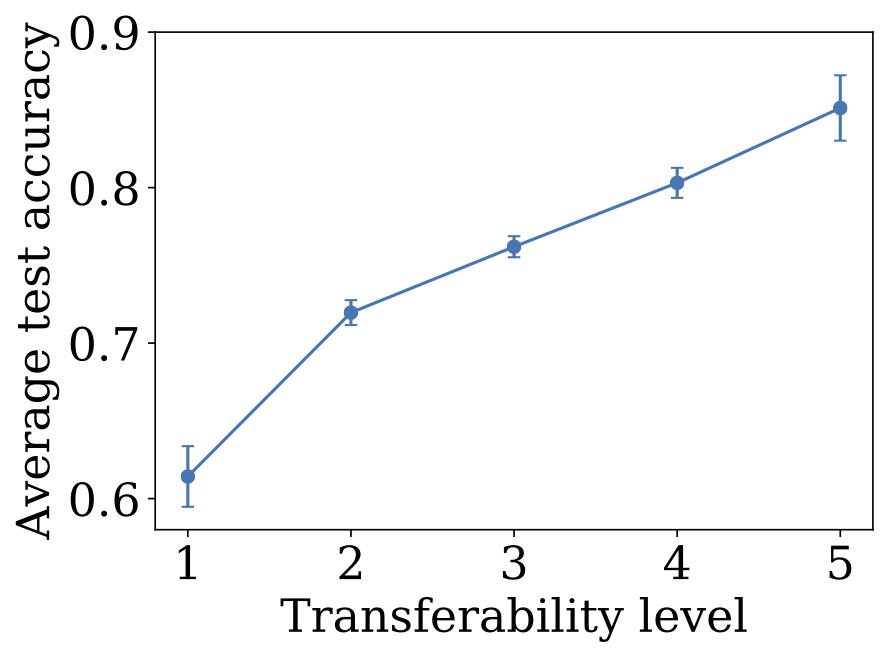}
}
\vskip -0.1in
\caption{{\bf Average test accuracy of CNAPs on tasks in five transferability levels} predicted from LEEP scores. The higher the level, the easier the transfer. See Sec.~\ref{sec:meta_transfer_exp} for details.}
\label{fig:cnaps}
\end{center}
\vskip -0.2in
\end{figure}

The last row of Table~\ref{tab:leep_compare} shows the correlation coefficient between LEEP scores and test accuracies of CNAPs. The coefficient is 0.591 with $p<0.001$, indicating that LEEP scores are a good measure of meta-transferability. Similar to Sec.~\ref{sec:small_data} and~\ref{sec:skewed_data}, Fig.~\ref{fig:cnaps} provides the average test accuracies of tasks in five LEEP score transferability levels. Fig.~\ref{fig:cnaps} clearly shows that higher transferability levels correspond to better test accuracies for CNAPs.

\subsection{LEEP vs.~Convergence of Fine-tuned Models}
\label{sec:convergence_exp}

For each target task, let us consider a \emph{reference} model: one that is trained from scratch using only the target data set. When it is easier to transfer from a source task to the target task, we often expect the fine-tuned models to converge more quickly and even exceed the performance of this reference model. The experiments in this section show that LEEP scores indeed predict this behavior.

We use the same small data settings defined in Sec.~\ref{sec:small_data} and train a reference model for each target task. Reference models are trained using SGD with 100 epochs for target tasks from CIFAR100 and with 40 epochs for those from FashionMNIST. When fine-tuning a model, we track the difference between the fine-tuned model's test accuracy and the \emph{final} test accuracy of the corresponding reference model (i.e., we always compare against the fully trained reference model). Similar to Sec.~\ref{sec:small_data}, we also consider different transferability levels according to LEEP scores, and average the accuracy differences of all tasks within the same transferability level.

\begin{figure}[t]
\begin{center}
\centerline{
\begin{subfigure}[t]{0.24\textwidth}
\includegraphics[width=\textwidth]{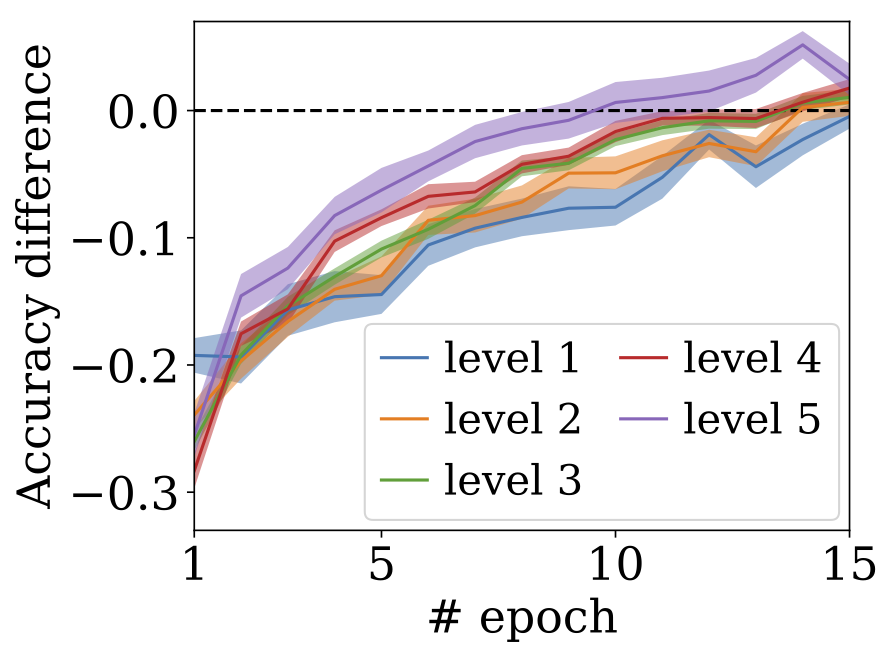}
\caption{ImageNet $\rightarrow$ CIFAR100}
\end{subfigure}
\begin{subfigure}[t]{0.24\textwidth}
\includegraphics[width=\textwidth]{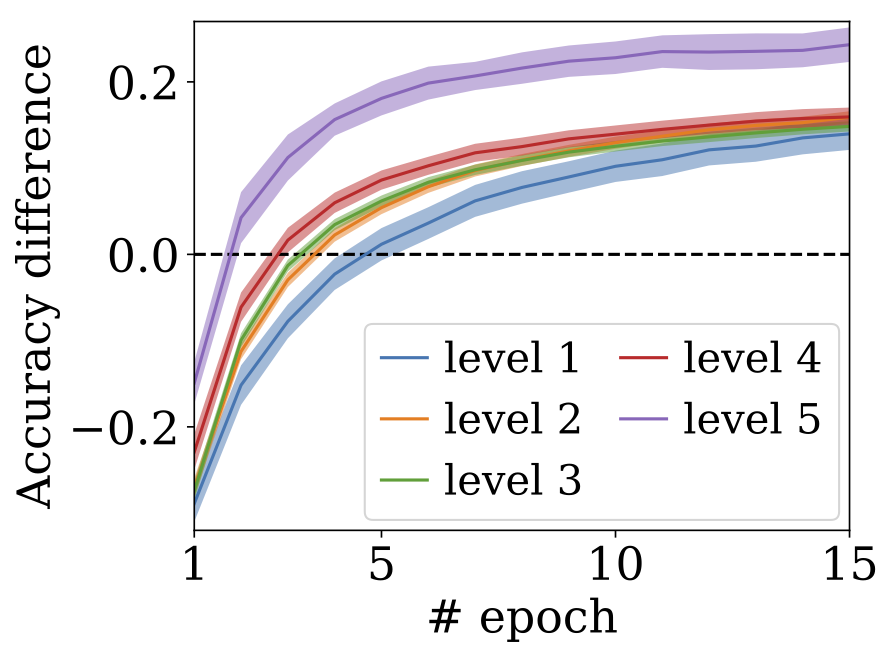}
\caption{CIFAR10 $\rightarrow$ CIFAR100}
\end{subfigure}
}
\caption{{\bf Convergence of accuracy of fine-tuned models} to the accuracy of \emph{reference} models trained from scratch using only the target data set. The convergence is represented by the accuracy difference between the fine-tuned and the reference models. Each curve is the average over tasks within the same transferability level. The zero lines indicate where the fine-tuned models match the accuracy of the reference models. See Sec.~\ref{sec:convergence_exp} for details.}
\label{fig:convergence}
\end{center}
\vskip -0.2in
\end{figure}

Fig.~\ref{fig:convergence} plots the average accuracy difference curves of five different transferability levels when transferring to CIFAR100 target tasks. Results for FashionMNIST target tasks are similar and given in Fig.~\ref{fig:convergence_full} in Appendix~\ref{sec:full_results}. From Fig.~\ref{fig:convergence}, on average, fine-tuned models on tasks in higher transferability levels have better convergence speeds (i.e., their curves reach zero faster). Furthermore, these models can outperform the reference models by larger margins (i.e., their curves reach higher values). In all cases, the fine-tuned models match the performance of the reference models using far fewer training epochs. These results confirm the advantage of transfer learning in small data settings, especially between highly transferable tasks, which in turn, can be efficiently predicted using our LEEP scores.

\subsection{Comparison of LEEP, NCE, and H scores}
\label{sec:compare_exp}

We compare the LEEP measure with the NCE measure proposed by~\citet{tran2019transferability} and the H score proposed by~\citet{bao2019information}. Particularly, for all the experimental settings in Sec.~\ref{sec:predict_exp},~\ref{sec:small_data},~\ref{sec:skewed_data}, and~\ref{sec:meta_transfer_exp} above, we compute the NCE score for each transfer from a source model to a target task using the method described in Sec.~\ref{sec:theory}. The computation is efficient since it also requires a single forward pass through the target data set to get the dummy labels. After obtaining the NCE scores, we evaluate their Pearson correlation coefficients and $p$ values with test accuracies (or F1 for imbalanced data) of the three transfer (or meta-transfer) learning algorithms. Similarly, we also compute the correlation coefficients and $p$ values of the H scores between the features returned by the source model and the target labels. The features are obtained by applying the source model to the target inputs.

Table~\ref{tab:leep_compare} shows the correlation coefficients of NCE and H scores in comparison with those of LEEP scores. When compared with NCE, LEEP scores have equal or better correlations with transfer performance in all except for two cases of the fine-tuning method (see the second and third rows of this method in Table~\ref{tab:leep_compare}). Even in these two cases, LEEP scores are only slightly worse than NCE scores. These comparisons confirm that LEEP scores are better than NCE scores for our transfer settings, with up to 30\% improvement in terms of correlation coefficients.

When compared with H scores, LEEP scores give better correlations in 16/23 cases. We note that the performance of H scores is not consistent in all experiments. Particularly, it completely fails to capture the transferability in 11 cases (those marked with asterisks). Even when it successfully captures transferability, it performs worse than LEEP on large target data sets. By comparison, our LEEP scores capture the transferability in all cases.

\begin{table*}[t]
\caption{{\bf Comparison of Pearson correlation coefficients} of LEEP, NCE~\citep{tran2019transferability}, and H scores~\citep{bao2019information}. Correlations are computed with respect to test accuracies (or F1) of three (meta)-transfer learning algorithms in various experimental settings. Correlations marked with asterisks (*) are not statistically significant ($p>0.05$), while the rest are statistically significant with $p<0.001$.}
\label{tab:leep_compare}
\vspace{-2mm}
\begin{center}
\resizebox{\textwidth}{!}{%
\begin{tabular}{c ccccc ccc}
\toprule
Algorithm \quad & \multicolumn{5}{c}{Experiment setting} & \multicolumn{3}{c}{Correlation coefficients} \\
\cmidrule(lr){2-6}
\cmidrule(lr){7-9}
& Source & Target & Source model & Properties of target data set & Details in & LEEP & NCE & H \\
\midrule
& CIFAR10 & CIFAR100 & ResNet20 & large, balanced & Sec.~\ref{sec:predict_exp} & {\bf 0.982} & {\bf 0.982} & 0.831 \\
& ImageNet & CIFAR100 & ResNet18 & large, balanced & Sec.~\ref{sec:predict_exp} & {\bf 0.974} & 0.973 & 0.924 \\
& CIFAR10 & CIFAR100 & ResNet20 & small, balanced & Sec.~\ref{sec:small_data} & 0.744 & 0.743 & {\bf 0.877} \\
& ImageNet & CIFAR100 & ResNet18 & small, balanced & Sec.~\ref{sec:small_data} & {\bf 0.798} & 0.715 & 0.026$^*$ \\
Re-train & CIFAR10 & FashionMNIST & ResNet20 & small, balanced & Sec.~\ref{sec:small_data} & 0.518 & 0.429 & {\bf 0.787} \\
head & ImageNet & FashionMNIST & ResNet18 & small, balanced & Sec.~\ref{sec:small_data} & {\bf 0.631} & 0.622 & 0.005$^*$ \\
& ImageNet & CIFAR100 & ResNet18 & small, balanced, noisy & Sec.~\ref{sec:small_data} & {\bf 0.612} & 0.579 & 0.017$^*$ \\
& CIFAR10 & CIFAR100 & ResNet20 & small, imbalanced & Sec.~\ref{sec:skewed_data} & {\bf 0.862} & 0.847 & 0.787 \\
& ImageNet & CIFAR100 & ResNet18 & small, imbalanced & Sec.~\ref{sec:skewed_data} & {\bf 0.522} & 0.484 & -0.058$^*$ \\
& CIFAR10 & FashionMNIST & ResNet20 & small, imbalanced & Sec.~\ref{sec:skewed_data} & 0.704 & 0.688 & {\bf 0.822} \\
& ImageNet & FashionMNIST & ResNet18 & small, imbalanced & Sec.~\ref{sec:skewed_data} & {\bf 0.645} & 0.624 & 0.059$^*$ \\
\midrule
& CIFAR10 & CIFAR100 & ResNet20 & large, balanced & Sec.~\ref{sec:predict_exp} & {\bf 0.967} & {\bf 0.967} & 0.787 \\
& ImageNet & CIFAR100 & ResNet18 & large, balanced & Sec.~\ref{sec:predict_exp} & 0.944 & {\bf 0.945} & 0.875 \\
& CIFAR10 & CIFAR100 & ResNet20 & small, balanced & Sec.~\ref{sec:small_data} & 0.396 & 0.401 & {\bf 0.737} \\
& ImageNet & CIFAR100 & ResNet18 & small, balanced & Sec.~\ref{sec:small_data} & {\bf 0.762} & 0.584 & -0.029$^*$ \\
Fine-tune & CIFAR10 & FashionMNIST & ResNet20 & small, balanced & Sec.~\ref{sec:small_data} & 0.339 & 0.258 & {\bf 0.826} \\
& ImageNet & FashionMNIST & ResNet18 & small, balanced & Sec.~\ref{sec:small_data} & {\bf 0.609} & 0.578 & 0.018$^*$ \\
& ImageNet & CIFAR100 & ResNet18 & small, balanced, noisy & Sec.~\ref{sec:small_data} & {\bf 0.348} & 0.324 & 0.06$^*$ \\
& CIFAR10 & CIFAR100 & ResNet20 & small, imbalanced & Sec.~\ref{sec:skewed_data} & 0.597 & 0.582 & {\bf 0.758} \\
& ImageNet & CIFAR100 & ResNet18 & small, imbalanced & Sec.~\ref{sec:skewed_data} & {\bf 0.565} & 0.503 & -0.069$^*$ \\
& CIFAR10 & FashionMNIST & ResNet20 & small, imbalanced & Sec.~\ref{sec:skewed_data} & 0.603 & 0.589 & {\bf 0.904} \\
& ImageNet & FashionMNIST & ResNet18 & small, imbalanced & Sec.~\ref{sec:skewed_data} & {\bf 0.507} & 0.425 & 0.056$^*$ \\
\midrule
CNAPS  & ImageNet & CIFAR100 & ResNet18 & small, balanced & Sec.~\ref{sec:meta_transfer_exp} & {\bf 0.591} & 0.310 & 0.025$^*$ \\
\bottomrule
\end{tabular}
}
\end{center}
\end{table*}

\subsection{LEEP for Source Model Selection}
\label{sec:model_selection}

We evaluate LEEP for the source model selection problem, where we need to select the best source model from 9 candidate models and transfer it to CIFAR100. The candidate models are pre-trained on ImageNet and  include ResNet18, ResNet34, ResNet50~\citep{he2016deep}, MobileNet1.0, MobileNet0.75, MobileNet0.5, MobileNet0.25~\citep{howard2017mobilenets}, DarkNet53~\citep{redmon2018yolov3}, and SENet154~\citep{hu2018squeeze}. Our target data set is the full CIFAR100 training set, while the target test set is the full CIFAR100 test set. We compare our LEEP measure to the NCE~\citep{tran2019transferability} and H score~\citep{bao2019information} baselines. We also consider ImageNet top-1 accuracy as an additional baseline since previous work~\citep{kornblith2019better} has shown that ImageNet accuracy can predict the performance of transferred models.

\begin{figure*}[!t]
\begin{center}
\includegraphics[width=\textwidth]{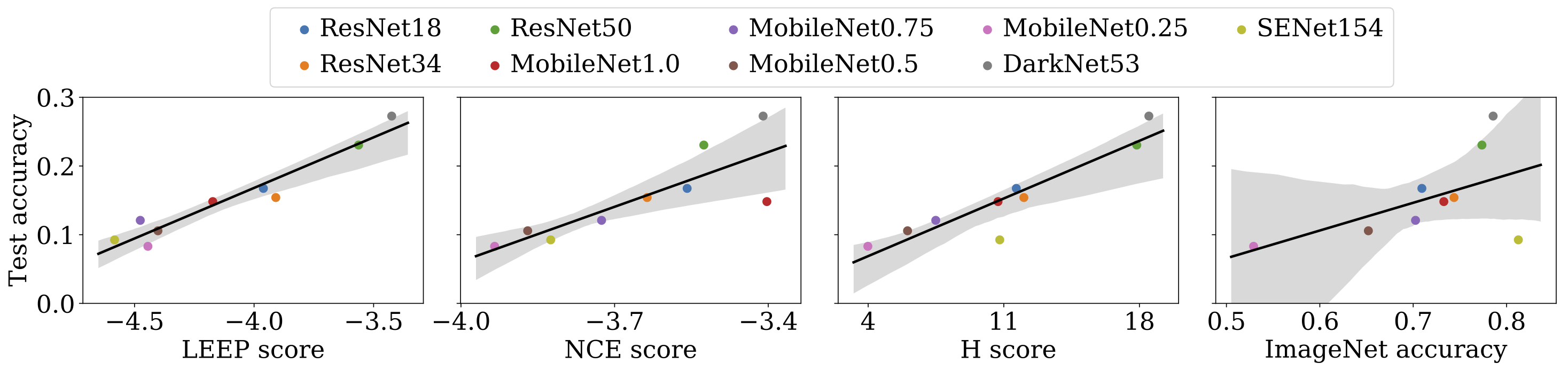}
\vskip -0.1in
\caption{{\bf Test accuracy vs.~transferability} predicted from LEEP, NCE~\citep{tran2019transferability}, H score~\citep{bao2019information}, and ImageNet top-1 accuracy~\citep{kornblith2019better} for 9 candidate source models (see the legend) pre-trained on ImageNet. The transferred models are obtained by re-training the head classifier. See Sec.~\ref{sec:model_selection} for details.}
\label{fig:model_select}
\end{center}
\end{figure*}

Fig.~\ref{fig:model_select} shows the results of this experiment for the head re-training method. From the figure, LEEP scores can predict well the test accuracy of models whose head classifiers are re-trained. In comparison, the other baselines all perform worse than LEEP. For example, NCE fails to predict the performance of MobileNet1.0, while H score and ImageNet accuracy fail on the SENet154. 

We also give results for the fine-tuning method in Fig.~\ref{fig:model_select_full} in Appendix~\ref{sec:full_results}. Generally, all the considered measures do not predict well the test accuracy of fine-tuned models, especially for ResNet18 and ResNet34 source models. One possible explanation is that the performance of fine-tuned models is sensitive to the architecture and the size of the source networks. However, the transferability measures considered in this section do not take these factors into account.

\section{Discussions}

We proposed LEEP, a novel transferability measure that can efficiently estimate the performance of transfer and meta-transfer learning algorithms before actually executing them. We show both theoretically and empirically that LEEP is an effective transferability measure that is useful in several scenarios. Below are more discussions about our work.

\minisection{Source model assumption}
Our work assumes a source model pre-trained on the source task. If, instead, we are given the source data set, we can first train a source model from the data and then use this model to compute the LEEP scores. The LEEP scores would depend on the architectural choice and thus the performance of the source model. This is expected and has been pointed out in previous work~\citep{kornblith2019better, tran2019transferability}.

\minisection{Rationale for LEEP and possible extensions}
The design of LEEP was aimed at being \emph{algorithm independence}, i.e., it should work for many different transfer learning algorithms. In fact, the performance of transfer learning algorithms depends partially on the relationship between the source model and the target data set, and thus LEEP tries to quantify such relationship to achieve this independence. We note that LEEP was not intended to be a transfer learning algorithm although the EEP can be viewed as a transfer learning algorithm that was designed for simplicity and speed instead of accuracy. Transfer learning should be performed by a proper algorithm, e.g., by re-training the head classifier or fine-tuning the model parameters.

Although it might seem surprising that LEEP works well in our experiments without explicitly using the feature distributions, we note that LEEP in fact takes into account the feature distributions \emph{indirectly}. Specifically, given a pre-trained source network, the output label distribution is a linear transformation of the features (i.e., output of the penultimate layer) followed by the Softmax. Thus, the dummy source label distribution indirectly contains information about the input features.

We can extend LEEP to include the learned features by transforming the feature vector into a probability distribution directly using a Softmax, and then computing LEEP with this new dummy distribution. In this case, the support of the dummy distribution has the same length as that of the feature vector and would have a different interpretation than ours. This would be a direction for future work.

\minisection{Effects of heterogeneous source and target tasks}
LEEP can be applied when the source and target tasks have different label semantics or when their input spaces are heterogeneous. In these cases, the source model would be more uncertain about the inputs from the target data set, leading to a more uniform dummy label distribution. By definition, LEEP scores will be smaller and thus indicating a harder transfer for these cases.

\minisection{Applications of LEEP}
Our experiments, reported in Sec.~\ref{sec:experiments}, showed that LEEP is applicable in diverse scenarios. In general, LEEP scores can be used to efficiently select highly transferable pairs of source and target tasks, yielding high transfer accuracy and good convergence speeds. This ability can support source model selection in transfer/meta-transfer learning scenarios, where we need to select a source model among several others in order to optimize for best performance on a target task.

Aside from transfer and meta-transfer learning, our LEEP scores are potentially useful for continual learning~\citep{zenke2017continual, swaroop2019improving}, multi-task learning~\citep{misra2016cross, standley2019tasks}, and feature selection~\citep{yosinski2014transferable}. For instance, LEEP scores can be used to estimate the hardness of task sequences, thereby helping to analyze properties of continual learning algorithms~\citep{nguyen2019toward}. For multi-task learning, LEEP scores can be used for selecting groups of tasks for joint training~\citep{standley2019tasks}. Finally, LEEP scores could be useful for hyperparameter transfer learning and Bayesian optimization~\citep{perrone2018scalable}. These are promising research directions that we leave to future work.

\appendix

\begin{figure*}[t]
\begin{center}
\centerline{
\begin{subfigure}[t]{0.25\textwidth}
\includegraphics[width=\textwidth]{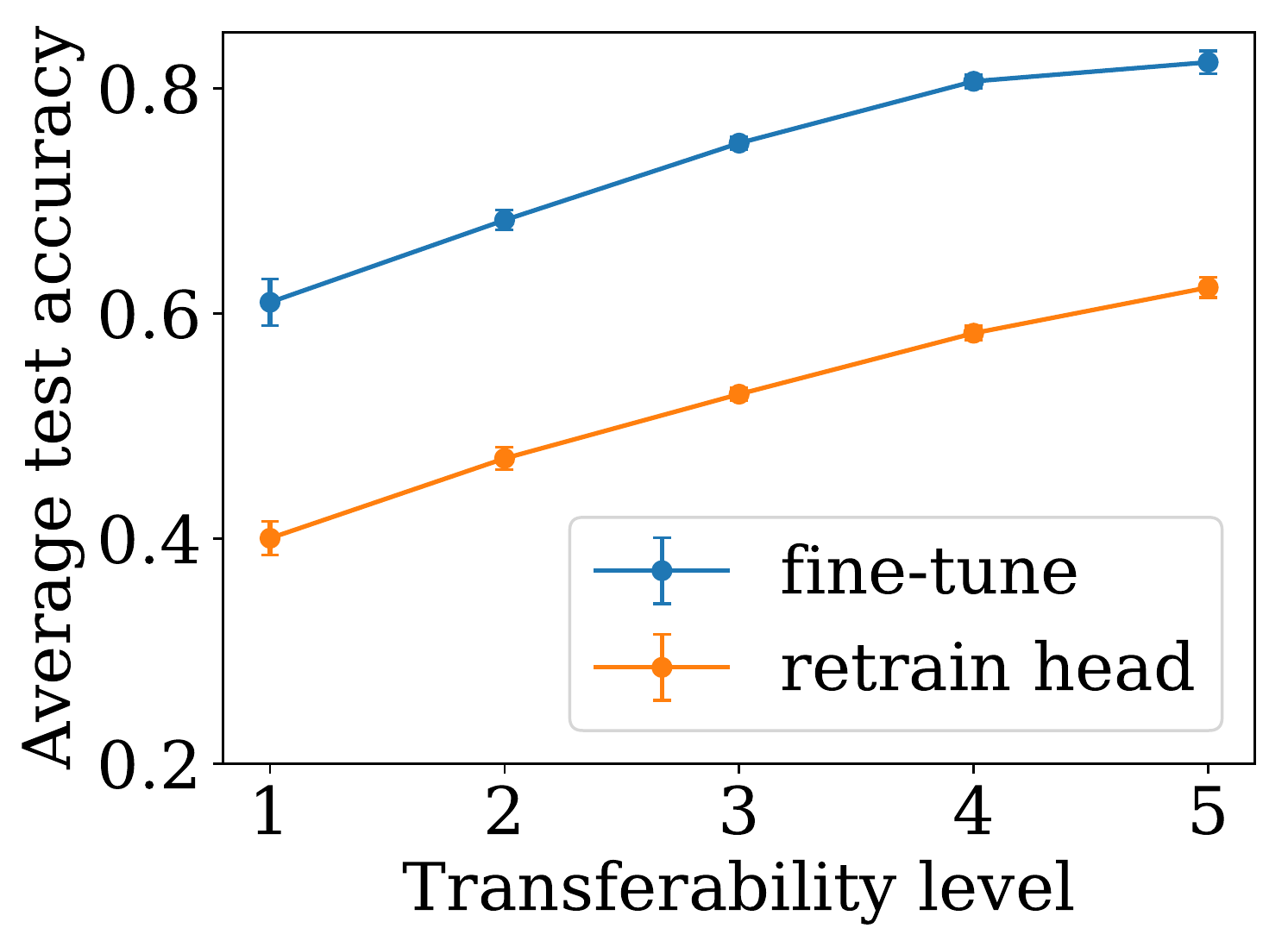}
\caption{ImageNet $\rightarrow$ CIFAR100}
\end{subfigure}
\begin{subfigure}[t]{0.25\textwidth}
\includegraphics[width=\textwidth]{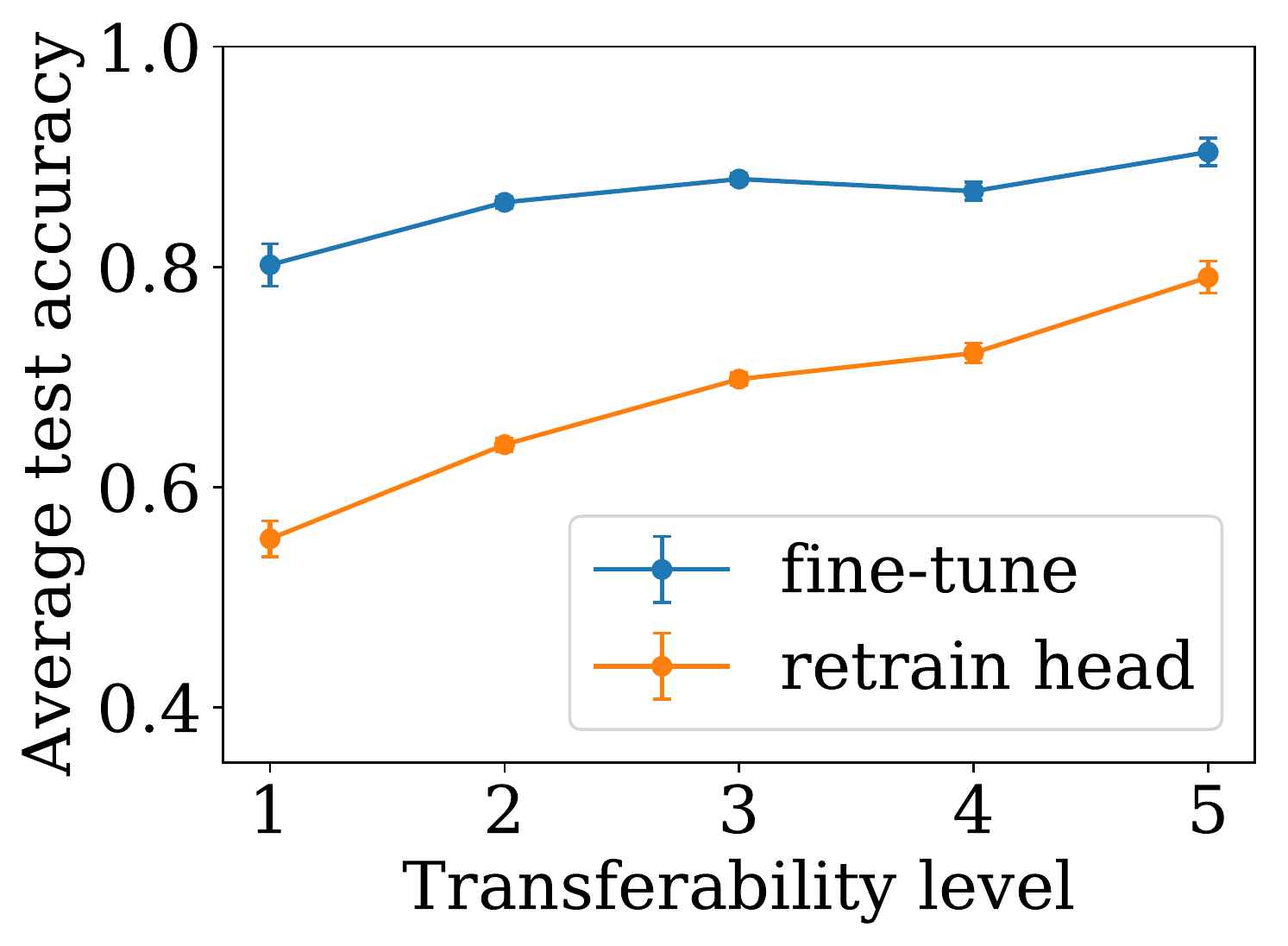}
\caption{CIFAR10 $\rightarrow$ CIFAR100}
\end{subfigure}
\begin{subfigure}[t]{0.25\textwidth}
\includegraphics[width=\textwidth]{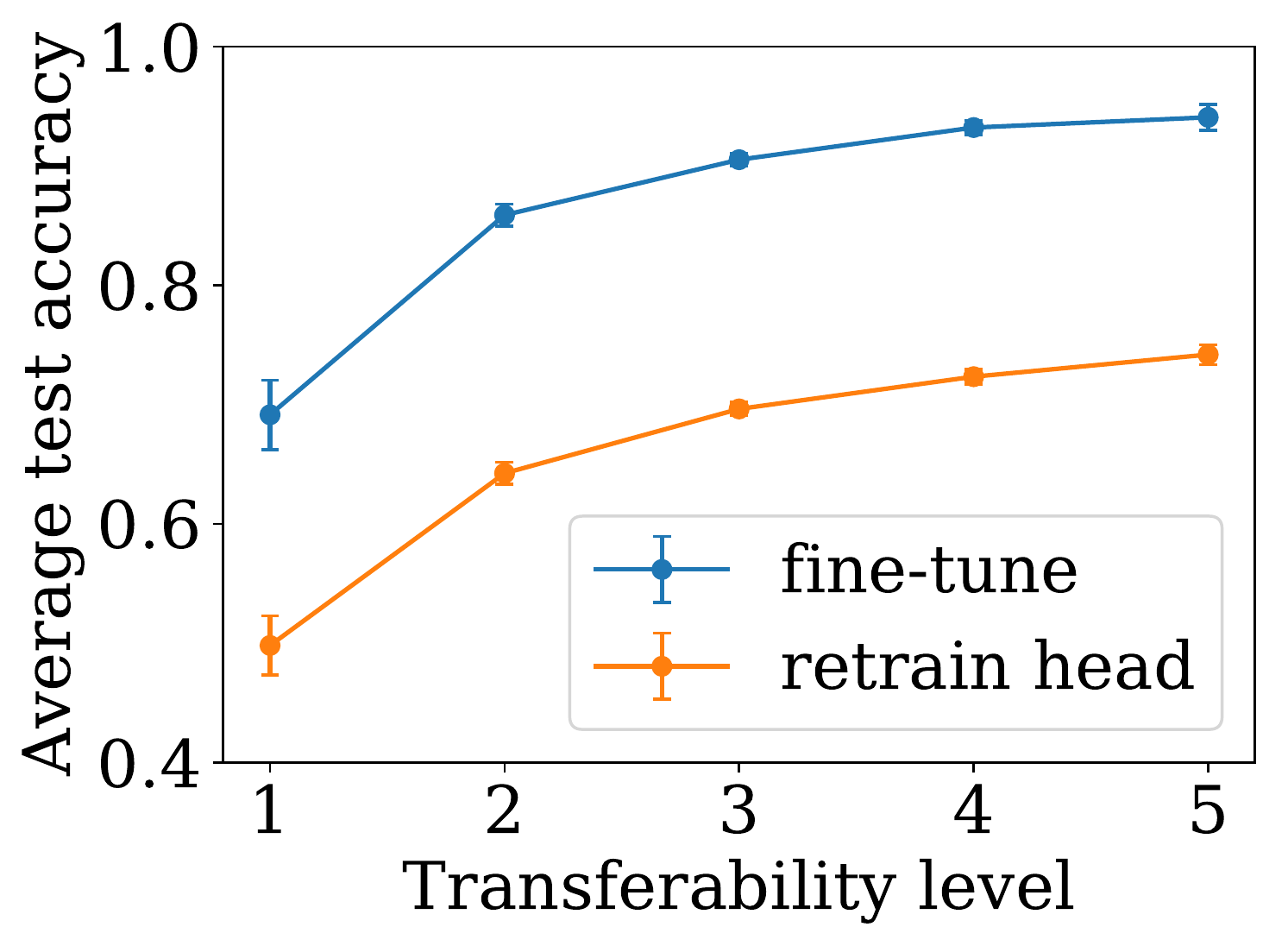}
\caption{ImageNet $\rightarrow$ FashionMNIST}
\end{subfigure}
\begin{subfigure}[t]{0.25\textwidth}
\includegraphics[width=\textwidth]{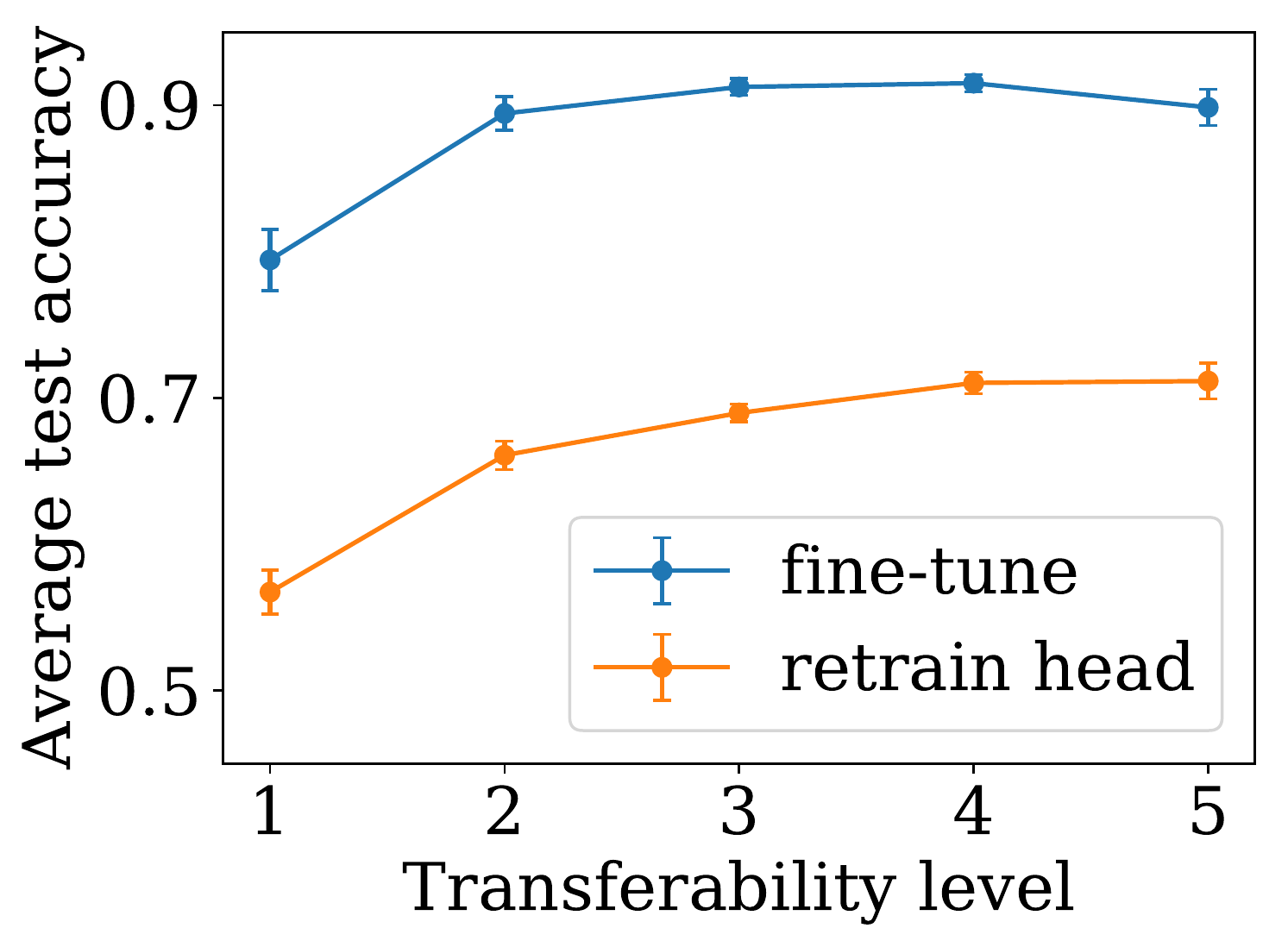}
\caption{CIFAR10 $\rightarrow$ FashionMNIST}
\end{subfigure}
}
\caption{{\bf Average test accuracy of transferred models on small, balanced target data sets in five transferability levels} obtained from LEEP scores. The higher the level, the easier the transfer. A $\rightarrow$ B in the subcaptions indicate that the source model is trained on A and the target datasets are constructed from B. The source models are ResNet18 for ImageNet (a,c) and ResNet20 for CIFAR10 (b,d).}
\label{fig:predict_small_full}
\end{center}
\end{figure*}

\begin{figure*}[t]
\begin{center}
\centerline{
\begin{subfigure}[t]{0.25\textwidth}
\includegraphics[width=\textwidth]{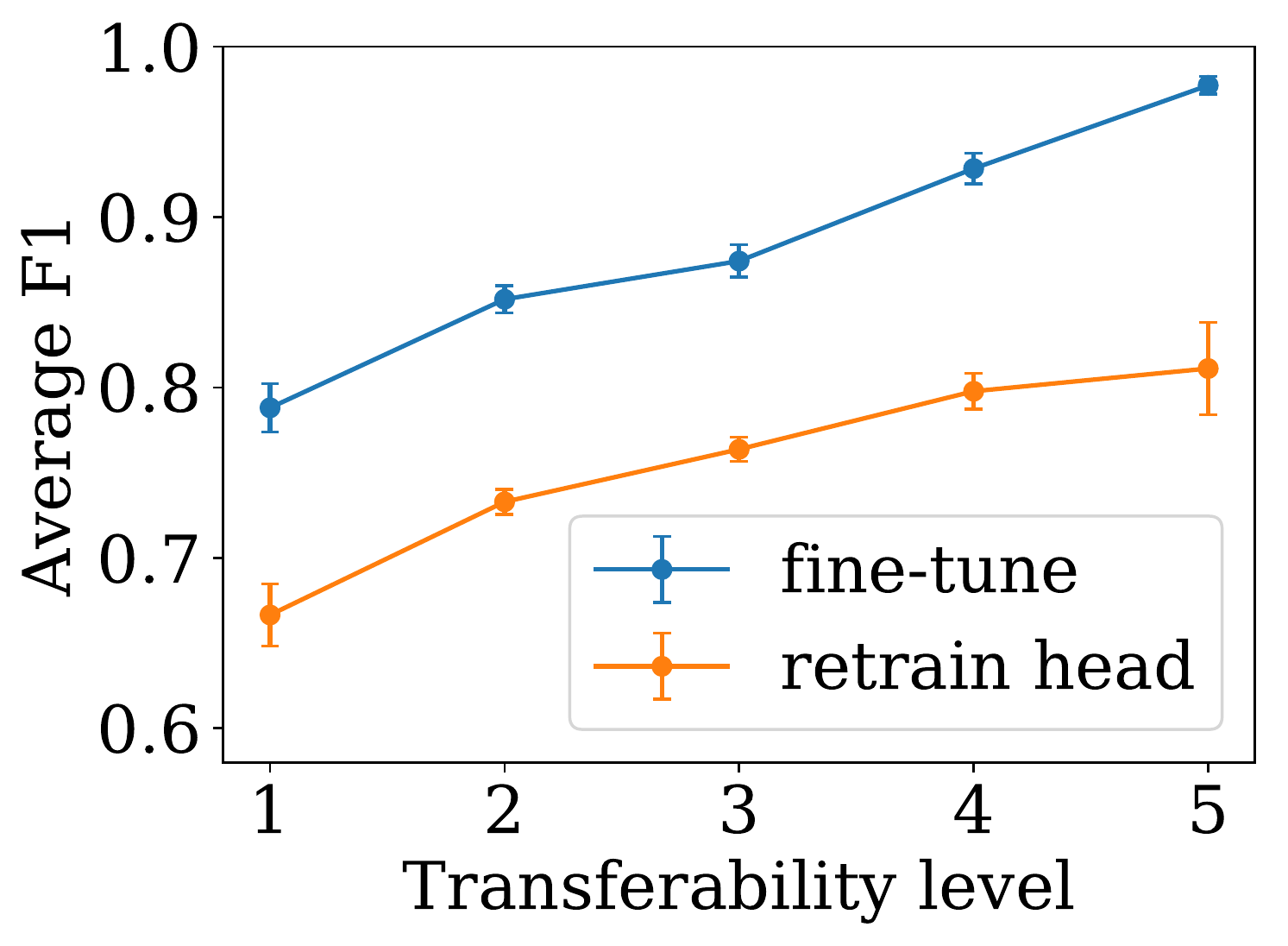}
\caption{ImageNet $\rightarrow$ CIFAR100}
\end{subfigure}
\begin{subfigure}[t]{0.25\textwidth}
\includegraphics[width=\textwidth]{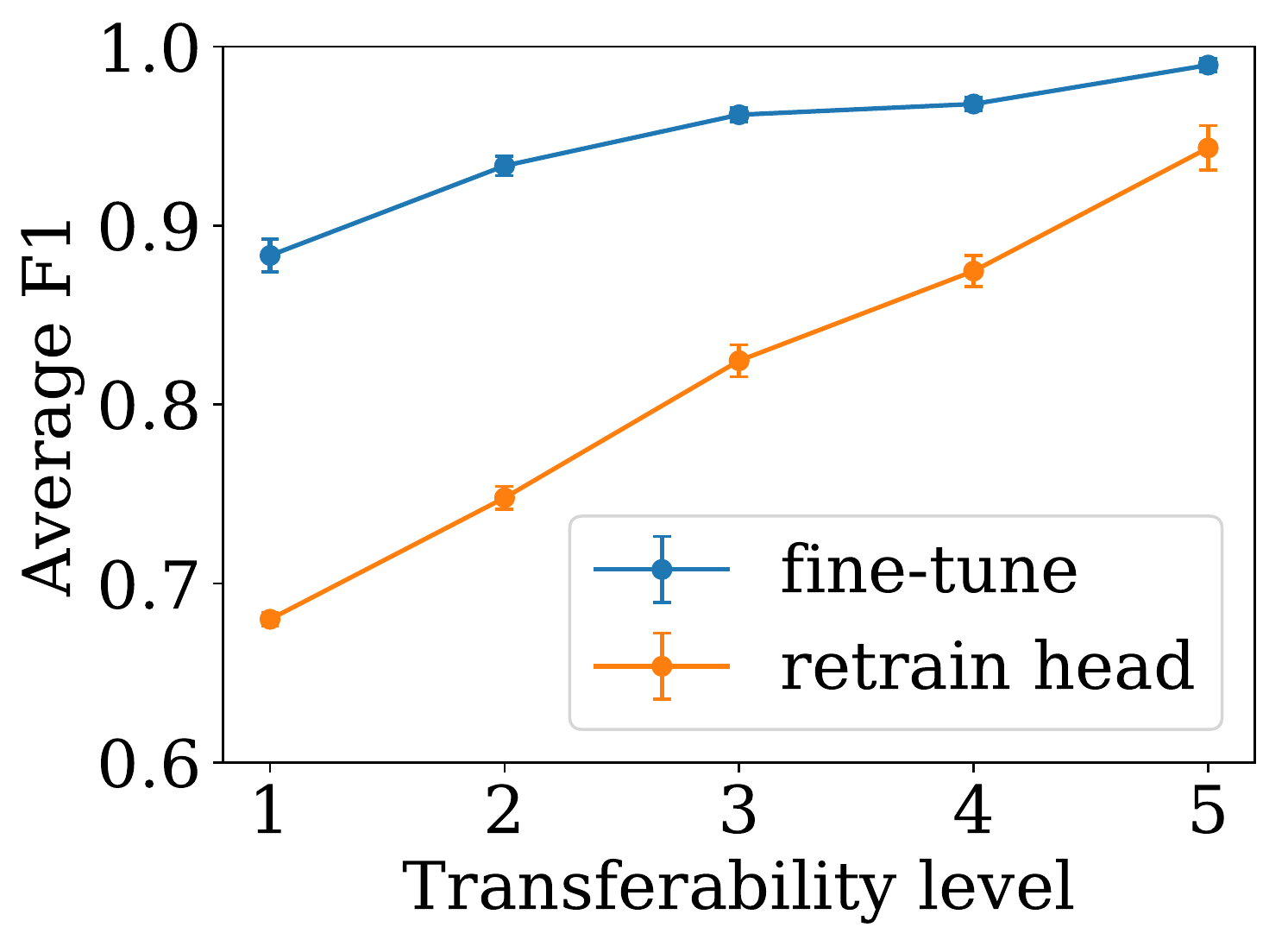}
\caption{CIFAR10 $\rightarrow$ CIFAR100}
\end{subfigure}
\begin{subfigure}[t]{0.25\textwidth}
\includegraphics[width=\textwidth]{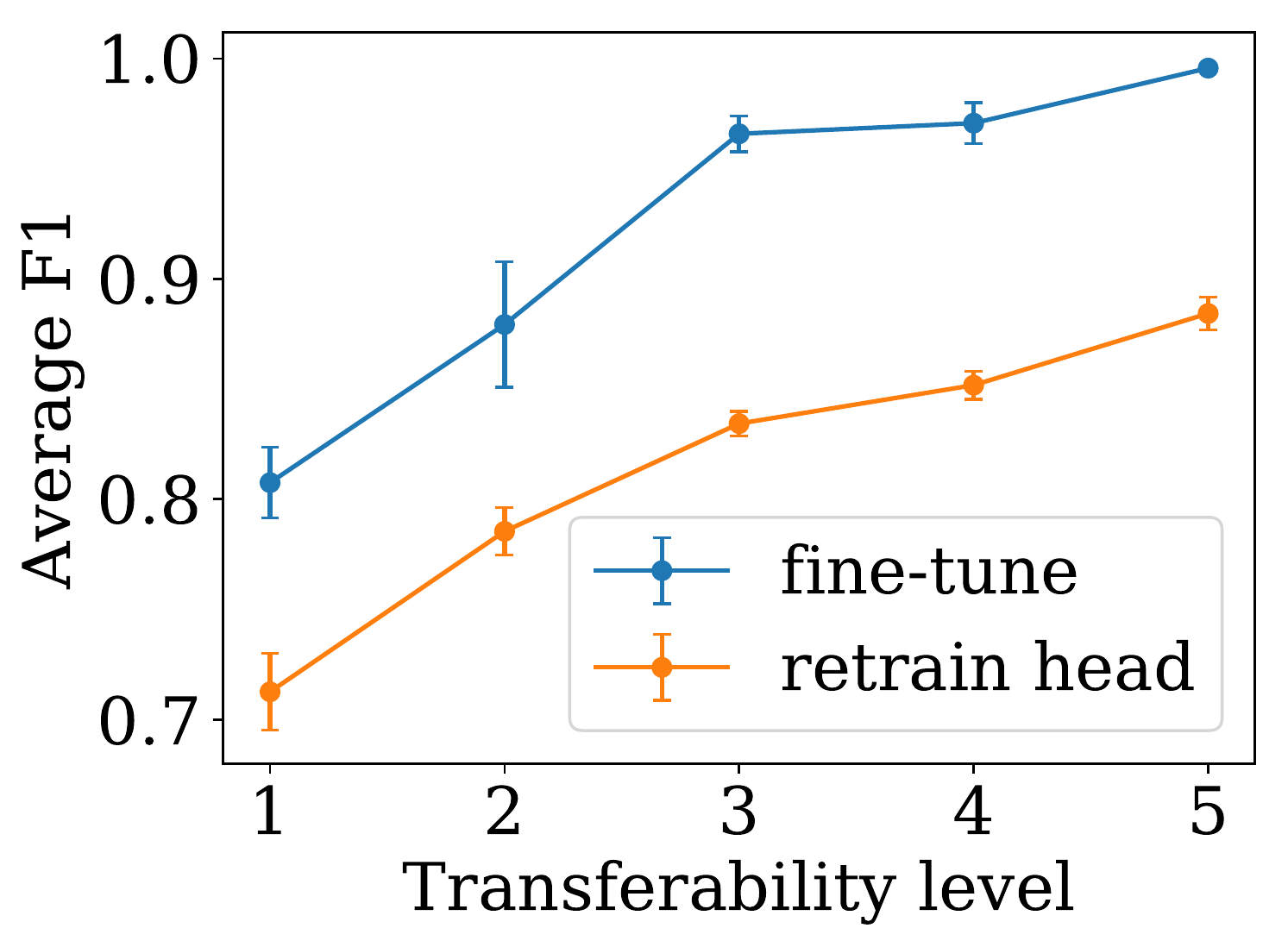}
\caption{ImageNet $\rightarrow$ FashionMNIST}
\end{subfigure}
\begin{subfigure}[t]{0.25\textwidth}
\includegraphics[width=\textwidth]{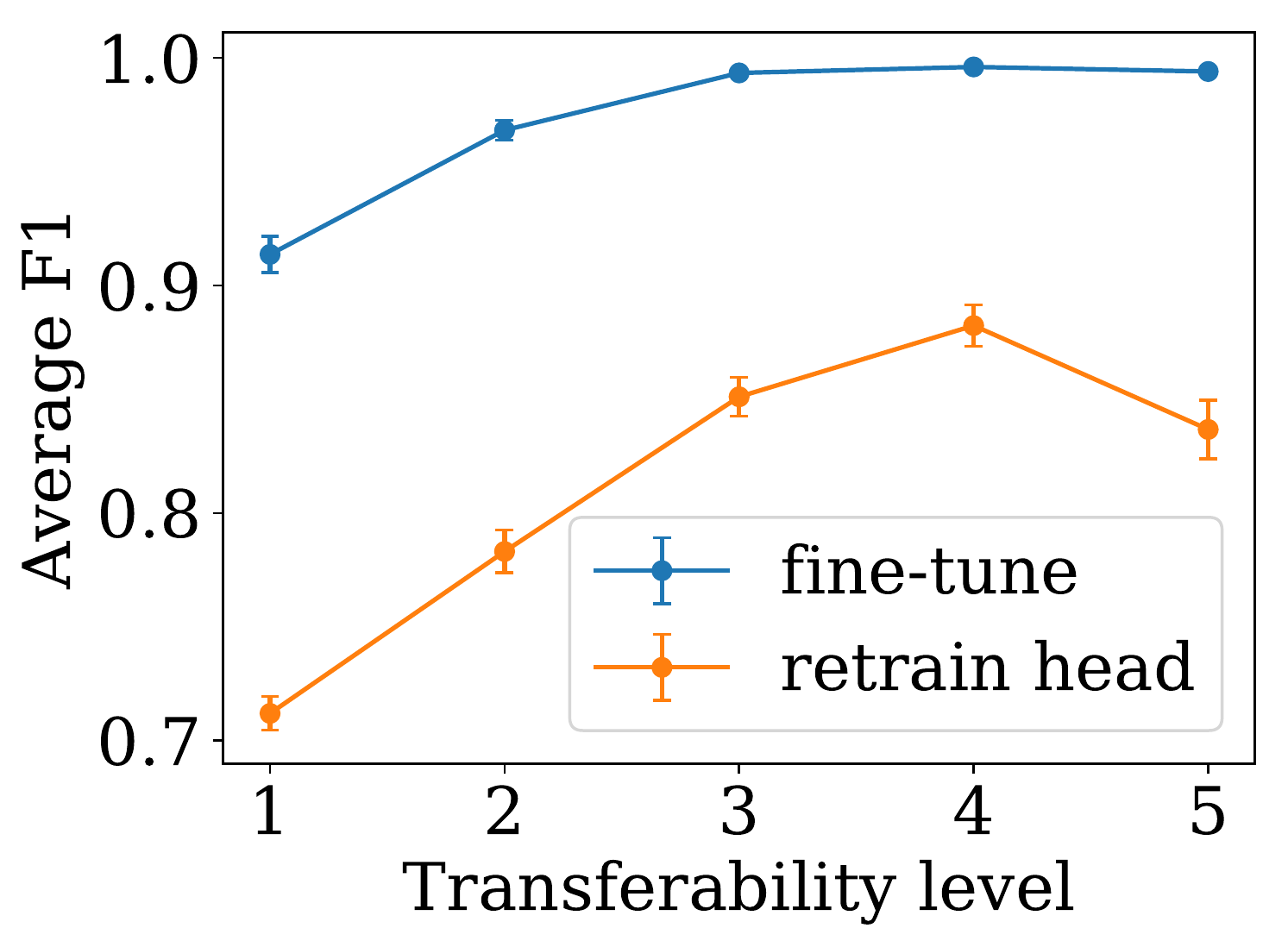}
\caption{CIFAR10 $\rightarrow$ FashionMNIST}
\end{subfigure}
}
\caption{{\bf Average test F1 score of transferred models on small, imbalanced target data sets in five transferability levels} obtained from LEEP scores. The higher the level, the easier the transfer. A $\rightarrow$ B in the subcaptions indicate that the source model is trained on A and the target datasets are constructed from B. The source models are ResNet18 for ImageNet (a,c) and ResNet20 for CIFAR10 (b,d).}
\label{fig:predict_skewed_full}
\end{center}
\end{figure*}

\section{Proofs}
\label{sec:proofs}

\subsection{Proof of Property~\ref{prop:upper_bound}}
This proof is straight-forward because $l(w, k^*)$ is the maximal average log-likelihood over $k \in \mathcal{K}$, $T(\theta, \mathcal{D})$ is the average log-likelihood of the EEP, and the EEP is in $\mathcal{K}$. Thus, $T(\theta, \mathcal{D}) \le l(w, k^*)$.

\subsection{Proof of Property~\ref{prop:lower_bound}}
Let $Z = (z_1, z_2, \ldots, z_n)$ be the dummy labels of $(x_1, x_2, \ldots, x_n)$ obtained when computing the NCE, and let $Y = (y_1, y_2, \ldots, y_n)$ be the true label set. We have:
\begin{align*}
T(\theta, \mathcal{D}) &= \frac{1}{n} \sum_{i=1}^n \log \left( \sum_{z \in \mathcal{Z}} \hat{P}(y_i | z) ~ \theta(x_i)_z \right) \tag{by definition} \\
&\ge \frac{1}{n} \sum_{i=1}^n \log \left( \hat{P}(y_i | z_i) ~ \theta(x_i)_{z_i} \right) \tag{monotonicity of log} \\
&= \frac{1}{n} \sum_{i=1}^n \log \hat{P}(y_i | z_i) +  \frac{1}{n} \sum_{i=1}^n \log \theta(x_i)_{z_i}.
\end{align*}
According to the proof of Theorem 1 of~\citet{tran2019transferability}, we have:
\begin{equation*}
\mathrm{NCE}(Y | Z) =  \frac{1}{n} \sum_{i=1}^n \log \hat{P}(y_i | z_i).
\end{equation*}
Thus, we have:
\begin{equation*}
T(\theta, \mathcal{D}) \ge \mathrm{NCE}(Y | Z) + \frac{1}{n} \sum_{i=1}^n \log \theta(x_i)_{z_i}.
\end{equation*}

\section{Full Experimental Results}
\label{sec:full_results}

Fig.~\ref{fig:predict_small_full} shows the results for all experimental settings with small balanced target data sets.

Fig.~\ref{fig:predict_skewed_full} shows the results for all experimental settings with small imbalanced target data sets.

\begin{figure*}[t]
\begin{center}
\centerline{
\begin{subfigure}[t]{0.25\textwidth}
\includegraphics[width=\textwidth]{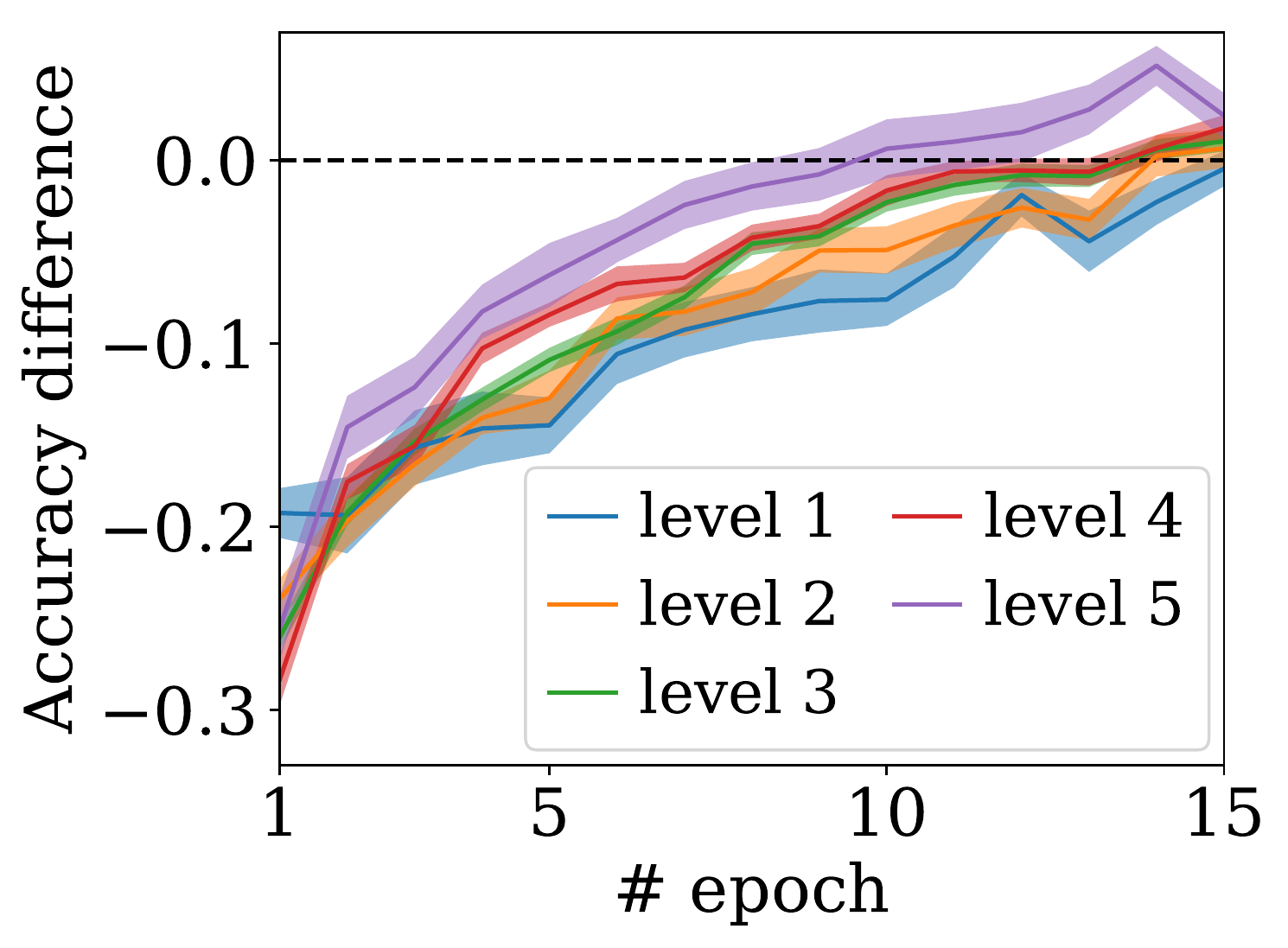}
\caption{ImageNet $\rightarrow$ CIFAR100}
\end{subfigure}
\begin{subfigure}[t]{0.25\textwidth}
\includegraphics[width=\textwidth]{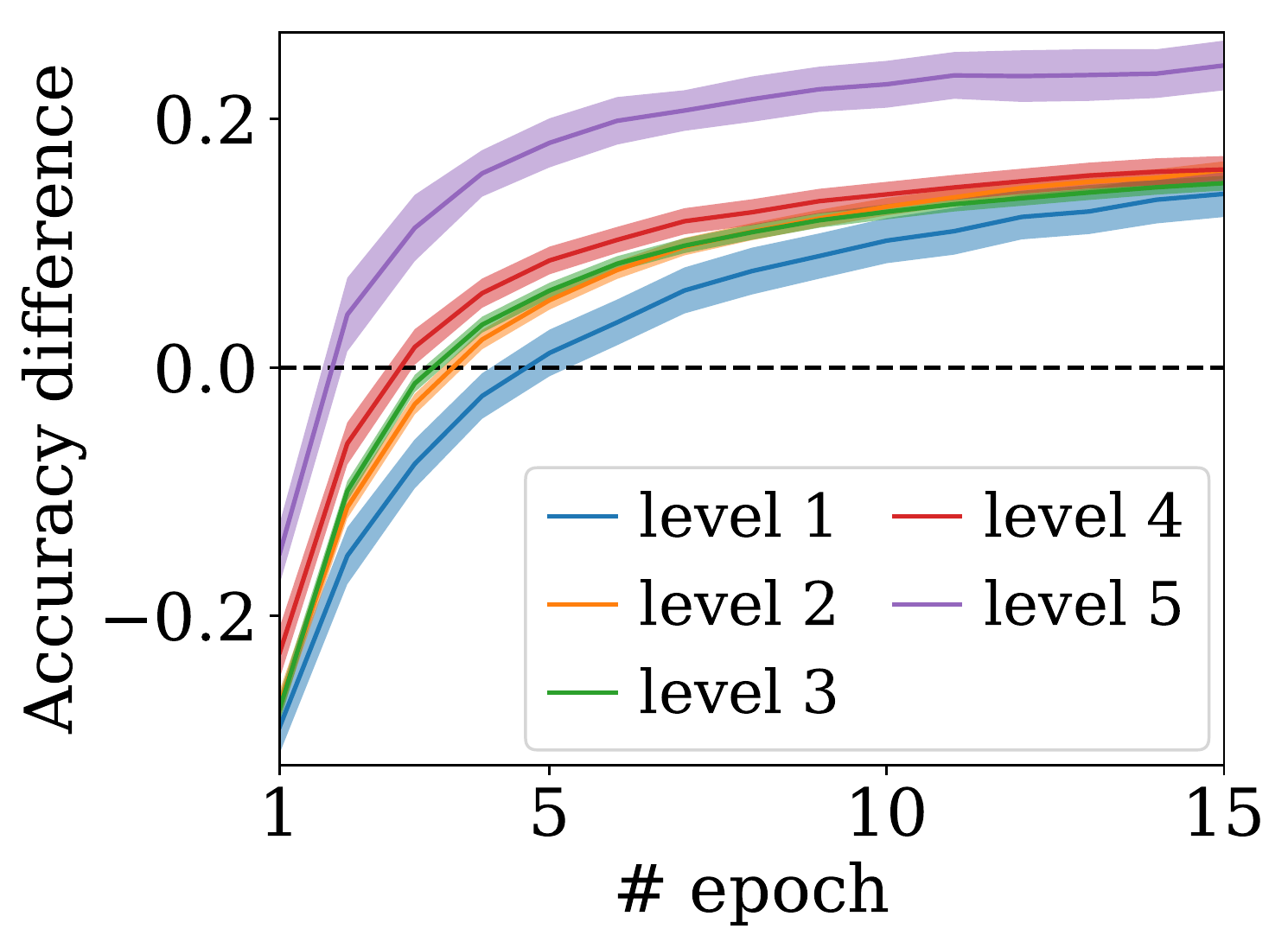}
\caption{CIFAR10 $\rightarrow$ CIFAR100}
\end{subfigure}
\begin{subfigure}[t]{0.25\textwidth}
\includegraphics[width=\textwidth]{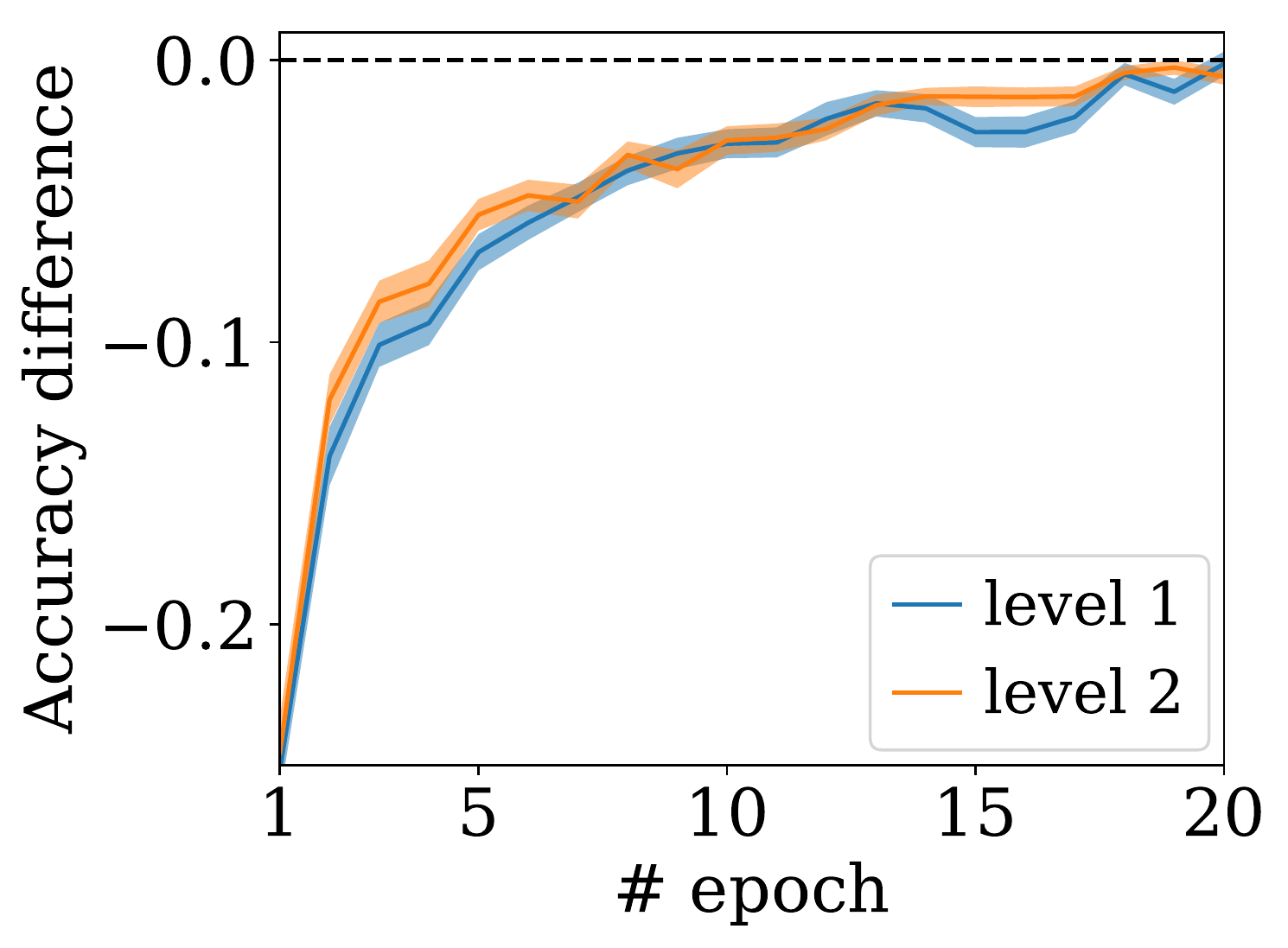}
\caption{ImageNet $\rightarrow$ FashionMNIST}
\end{subfigure}
\begin{subfigure}[t]{0.25\textwidth}
\includegraphics[width=\textwidth]{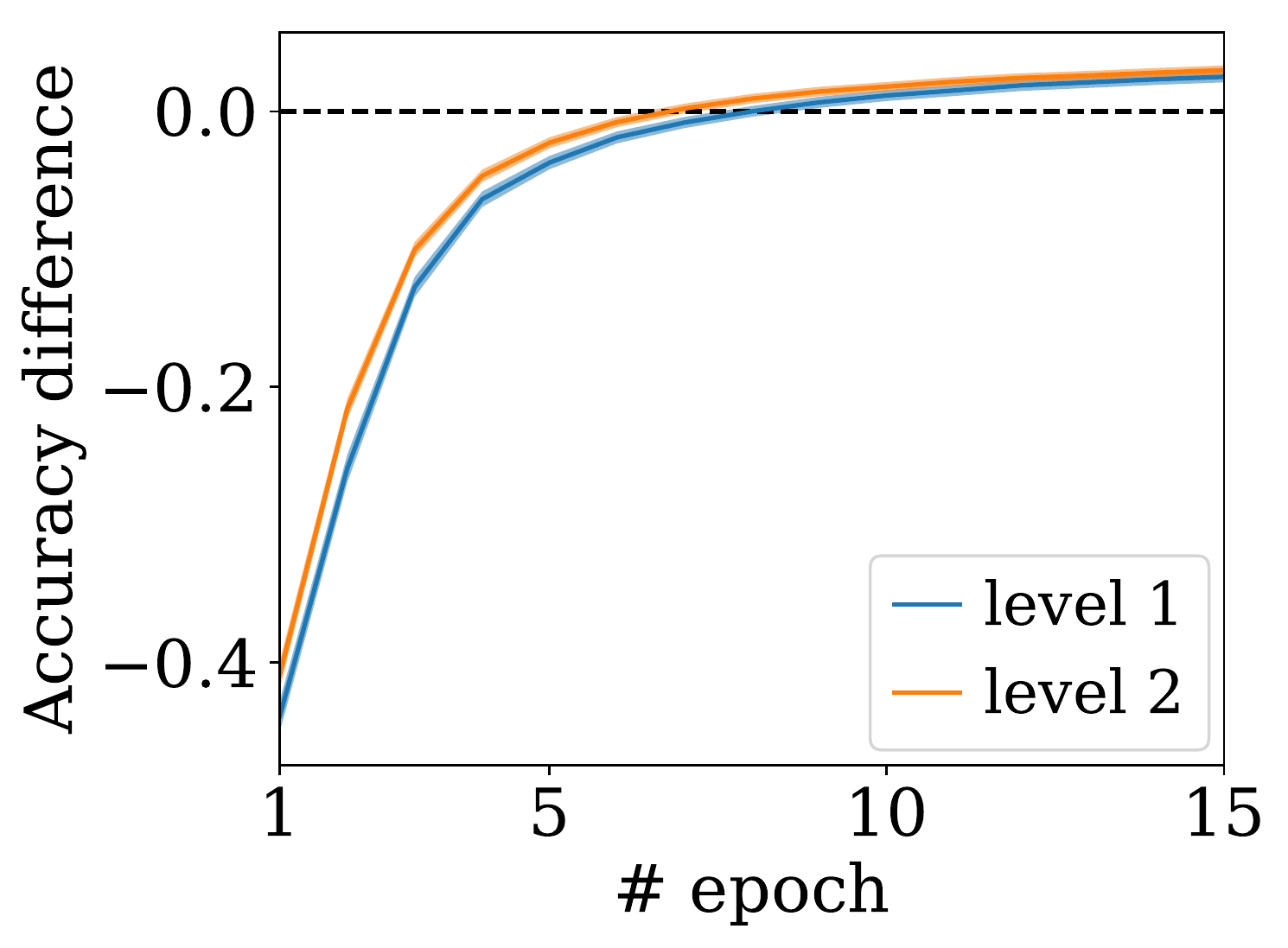}
\caption{CIFAR10 $\rightarrow$ FashionMNIST}
\end{subfigure}
}
\caption{{\bf Convergence of accuracy for fine-tuned models} to the accuracy of a \emph{reference} model trained from scratch using only the target dataset. The convergence is represented by the accuracy difference between the fine-tune model and the reference model. Each curve is the average of the accuracy difference curves over tasks within the same transferability level. The zero lines indicate where the fine-tuned models match the accuracy of the reference model.}
\label{fig:convergence_full}
\end{center}
\end{figure*}

Fig.~\ref{fig:convergence_full} shows the results for all experimental settings with the convergence speed of fine-tuned models. For a clearer comparison, we only consider two LEEP transferability levels for target tasks constructed from FashionMNIST.

\begin{figure*}[!t]
\begin{center}
\begin{subfigure}[b]{\textwidth}
\includegraphics[width=\textwidth]{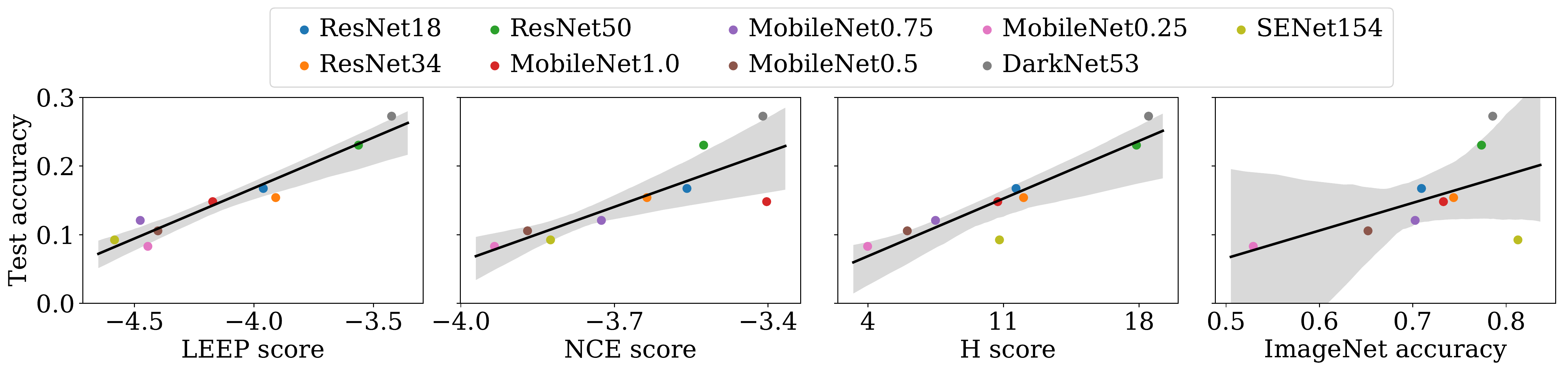}
\vskip -0.05in
\caption{Re-train head}
\end{subfigure}

\vskip 0.1in

\begin{subfigure}[b]{\textwidth}
\includegraphics[width=\textwidth]{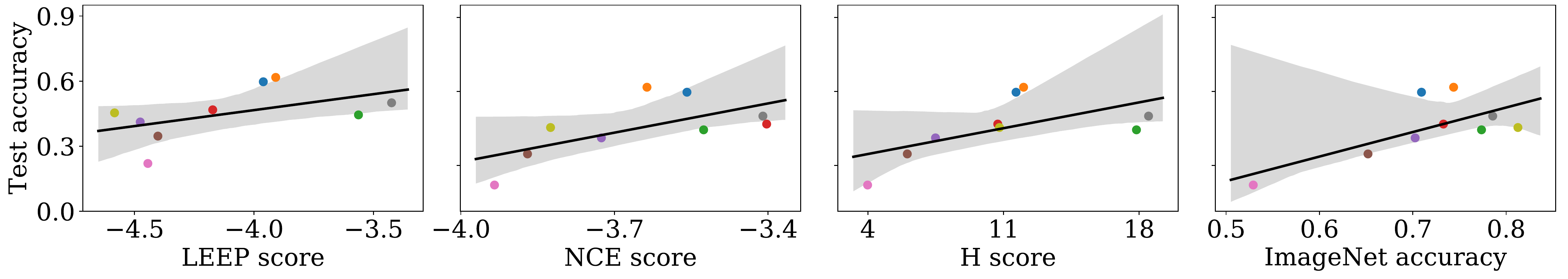}
\vskip -0.05in
\caption{Fine-tune}
\end{subfigure}
\vskip -0.05in
\caption{{\bf Test accuracy vs.~transferability} according to LEEP score, NCE score~\citep{tran2019transferability}, H score~\citep{bao2019information}, and ImageNet accuracy~\citep{kornblith2019better} for 9 candidate source models (see the legend) pre-trained on ImageNet. The transferred models are obtained by (a) re-training the head classifier, and (b) fine-tuning the source model.}
\label{fig:model_select_full}
\end{center}
\end{figure*}

Fig.~\ref{fig:model_select_full} shows the results for all experimental settings in the source model selection problem.

\newpage

\bibliography{leep}
\bibliographystyle{icml2020}

\end{document}